\documentclass[a4paper,fleqn]{cas-dc}
\usepackage[compress]{natbib}

\usepackage{pifont}
\usepackage{caption}
\usepackage{subfig}

\usepackage{times}
\usepackage{epsfig}
\usepackage{graphicx}
\usepackage{amsmath}
\usepackage{amssymb}
\usepackage{booktabs}
\usepackage{multirow}
\usepackage{pifont}
\usepackage{booktabs} % for professional tables
\usepackage{multirow}
\usepackage{svg}
\usepackage{xcolor} % For color
\usepackage{amsmath}
\usepackage[linesnumbered,ruled,vlined]{algorithm2e}
\usepackage{pifont}\newcommand{\cmark}{\ding{51}}%
\newcommand{\xmark}{\ding{55}}%
\usepackage{algpseudocode}
\usepackage{newunicodechar}
\newunicodechar{≥}{\ensuremath{\ge}}
\usepackage{float}
\usepackage{comment}
\usepackage{multirow}
\usepackage{rotating} % for rotatebox
\usepackage{array}

\usepackage{adjustbox}

\usepackage{xcolor}

\def\tsc#1{\csdef{#1}{\textsc{\lowercase{#1}}\xspace}}
\tsc{WGM}
\tsc{QE}
\tsc{EP}
\tsc{PMS}
\tsc{BEC}
\tsc{DE}
\begin{document}

\renewcommand{\floatpagefraction}{.8}
\renewcommand{\textfraction}{.1}
\renewcommand{\topfraction}{.9}
\renewcommand{\bottomfraction}{.8}

\shorttitle{RobustSurg: Endoscopic DG for surgical scene segmentation}

\shortauthors{M. Ali et~al.}

\title [mode = title]{
RobustSurg: Tackling domain generalisation for out-of-distribution surgical scene segmentation
}    
 
\author[inst1]{Mansoor Ali}
\author[inst2]{Maksim Richards}
\author[inst1]{Gilberto Ochoa-Ruiz*}
\author[inst3]{Sharib Ali*}

\affiliation[inst1]{organization={Escuela de Ingenieria y Ciencias, Tecnologico de Monterrey},%Department and Organization
            city={Monterrey},
            postcode={64849}, 
            state={N.L.},
            country={Mexico}}
            
\affiliation[inst2]{organization={St John’s College, University of Oxford},%Department and Organization
            city={Oxford},
            postcode={OX1 3JP}, 
            % state={Aguascalientes},
            country={UK}}

\affiliation[inst3]{organization={School of Computer Science, University of Leeds},%Department and Organization
            % addressline={Vandœuvre-les-Nancy cedex}, 
            city={Leeds},
            postcode={LS2 9JT}, 
            % state={Nancy},
            country={UK}}

\cortext[cor]{Corresponding author(s): \\  Gilberto Ochoa-Ruiz (gilberto.ochoa@tec.mx) \\ Sharib Ali ( s.s.ali@leeds.ac.uk)}

\maketitle
%%%%%%%%%%%%%%%%%%%%%%%%%%%%%%%%%%%%%%%%%%%%%%%%%%%%%%%%%%%%%%%%%%%%%%%%%%%%%%%%%%%%%%%%%%%%%
%%%%%%%%%%%%%%%%%%%%%%%%%%%%%%%%%%%%%%%%%%%%%%%%%%%%%%%%%%%%%%%%%%%%%%%%%%%%%%%%%%%%%%%%%%%%%
% TODO: MAKe sure that everywhere it is British English as your start suggests so...
\begin{abstract}
While recent advances in deep learning (DL) for surgical scene segmentation have demonstrated promising results on single-centre and single-imaging modality data, these methods usually do not generalise to unseen distribution (i.e., from other centres) and unseen modalities. Even though human experts can distinguish visual appearances, DL methods often fail to do so if data samples do not follow the independent and identically distributed (IID) data property, where IID refers to the data samples drawn from the same distribution. Current literature for tackling generalisation on out-of-distribution (OOD) data and domain gaps due to modality changes has been widely researched but mostly for natural scene data. However, these methods cannot be directly applied to the surgical scenes due to limited visual cues and often extremely diverse scenarios compared to the natural scene data. Inspired by these works in natural scene to push generalisability on OOD data, we hypothesise that exploiting the style and content information in the surgical scenes could minimise the appearances making it less variable to sudden changes such as blood or imaging artefacts. This can be achieved by performing instance normalisation and feature covariance mapping techniques for robust and generalisable feature representations. Further, to eliminate the risk of removing salient feature representation associated with the objects of interest, we introduce a restitution module within the feature learning ResNet backbone that can enable the retention of useful task-relevant features. To tackle lack of multiclass and multicentre data for surgical scene segmentation, we also provide a newly curated dataset that can be vital for addressing generalisability in this domain. Our proposed RobustSurg obtained nearly 23\% improvement on the baseline DeepLabv3+ and from 10-32\% improvement on the SOTA in terms of mean intersection over union (IoU) score on an unseen centre HeiCholSeg dataset when trained on CholecSeg8K. Similarly, RobustSurg also obtained nearly 22\% improvement over the baseline and nearly 11\% improvement on a recent state-of-the-art (SOTA) method for the target (different modality) set of EndoUDA polyp dataset. 
%EndoUDA or other data should not be your focus!
% Similarly, when trained on the CholecSeg8K and tested on the unseen centre HeiChole benchmark dataset, our method obtained nearly 4\% improvement on the baseline and from 1.9-6.3\% improvement on the SOTA.
\end{abstract}
%%%%%%%%%%%%%%%%%%%%%%%%%%%%%%%%%%%%%%%%%%%%%%%%%%%%%%%%%%%%%%%%%%%%%%%%%%%%%%%%%%%%%%%%%%%%%
%%%%%%%%%%%%%%%%%%%%%%%%%%%%%%%%%%%%%%%%%%%%%%%%%%%%%%%%%%%%%%%%%%%%%%%%%%%%%%%%%%%%%%%%%%%%%
% Keywords
% Each keyword is separated by \sep

%%%%%%%%%%%%%%%%%%%%%%%%%%%%%%%%%%%%%%%%%%%%%%%%%%%%%%%%%%%%%%%%%%%%%%%%%%%%%%%%%%%%%%%%%%%%%
%%%%%%%%%%%%%%%%%%%%%%%%%%%%%%%%%%%%%%%%%%%%%%%%%%%%%%%%%%%%%%%%%%%%%%%%%%%%%%%%%%%%%%%%%%%%%
%\linenumbers


\section{Introduction}
\label{sec:introduction}

%\subsection{General Context}
Minimally invasive surgery (MIS) has brought about a paradigm shift in modern surgical practice by reducing hospitalisation time, operative trauma, visible scars, and the risk of comorbidity \citep{buia2015laparoscopic}. However, the quality of a surgical procedure in the traditional operating room still largely depends on the surgical proficiency and experience of the surgeon. Moreover, the conditions and safety requirements during surgery can be quite demanding for the persons involved. Thus, AI tools have been developed to assist in interventions for tasks such as surgical tool navigation \citep{xu2023information}, skill assessment \citep{igaki2023automatic}, workflow analysis \citep{zhang2023surgical}, and action recognition \citep{sharma2023rendezvous}, among others.
% \begin{figure*}[hbt!]
%     \centering
% \includegraphics[width=0.9\linewidth]{img/DG_test_diagram_BE.pdf}
%     \caption{.}
%     \label{fig:qualitative}
% \end{figure*}

% \begin{figure*}[hbt!]
%     \centering
% \includegraphics[width=0.7\linewidth]{img/DG_test_v2.pdf}
%     \caption{.}
%     \label{fig:qualitative}
% \end{figure*}

\begin{figure*}[htbp!]
    \centering
\includegraphics[width=0.7\linewidth]{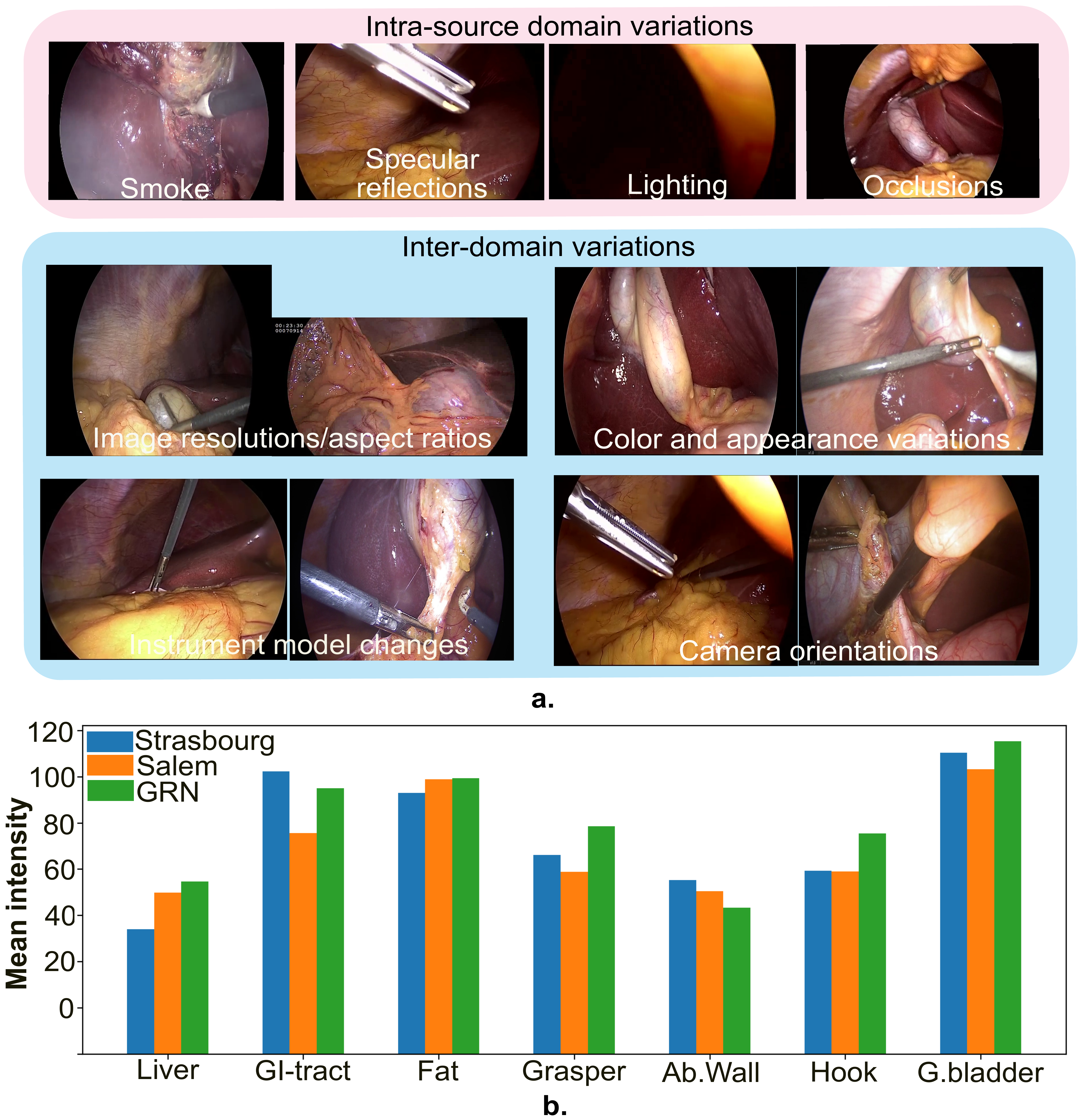}
    \caption{Overview of domain shift problem in the MIS. \textbf{a.} Top section shows the intra-source domain variations which we aim to exploit in learning a generalisable model, while bottom section shows shift problem between source and target domains. \textbf{b.} Comparison of intensity histograms between training source domain and different centre target domains.  }
    \label{fig:domain shift}
\end{figure*}

Image-guided procedures such as laparoscopic, endoscopic and radiological interventions heavily rely on images for surgical guidance and decision support. With the global expansion of minimally-invasive surgical practice and the number of surgeries performed, the volume of acquired video data is increasing rapidly. The recent success of deep learning approaches are the perfect avenues to use this data to further help clinicians in decision making, thus improving patient safety and healthcare. 

\begin{figure*}[t!]
    \centering
\includegraphics[width=1\linewidth]{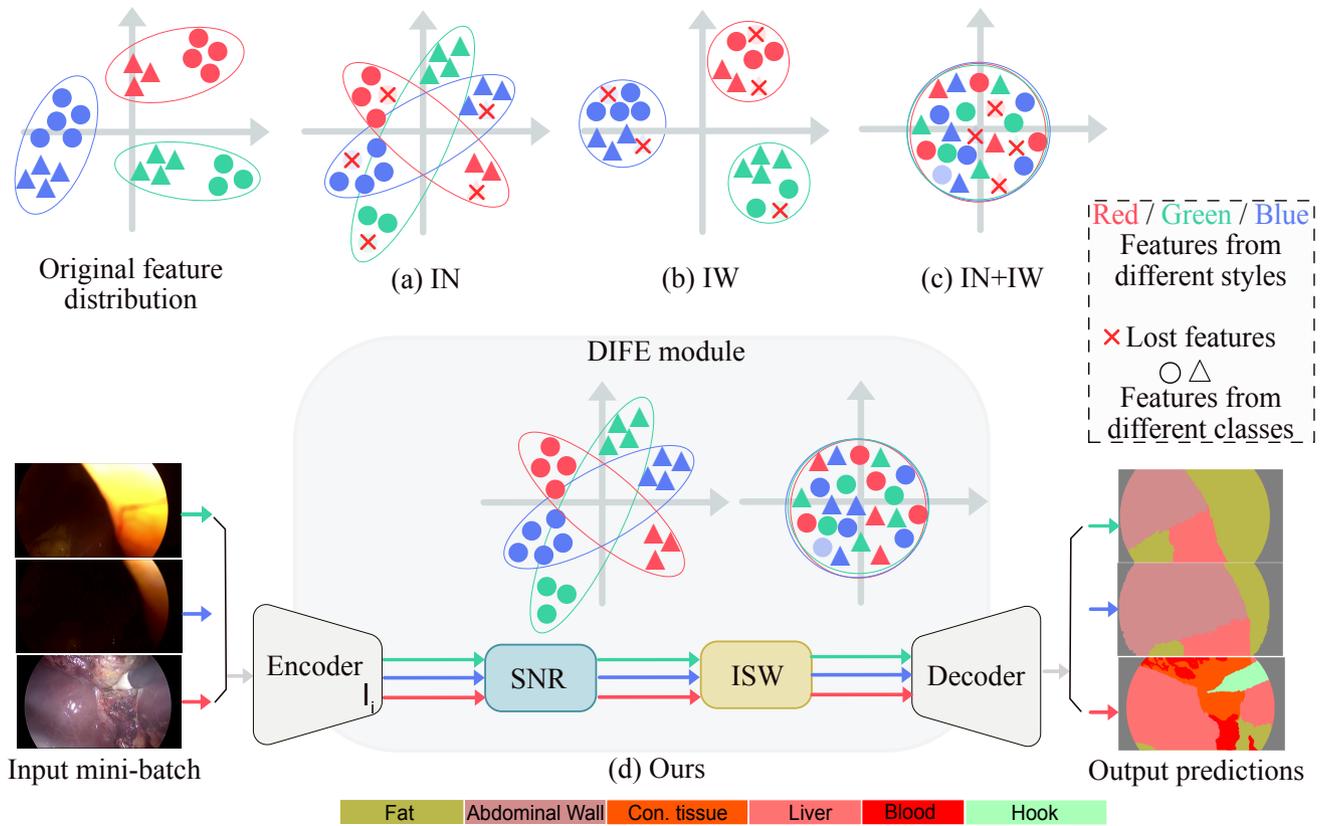}
    \caption{Comparative illustration of existing DG approaches and our proposed approach. Current Instance normalization (IN) and Instance whitening (IW) techniques (a-c) performs feature standardization and minimise global distribution variance but also loose some useful discriminative information (shown with faded symbols). Our proposed Domain-invariant feature encoder (DIFE) block aims to restore the lost features during IN stage. Note that this figure shows features from two classes for simplicity, but the idea applies to all classes.}
    \label{fig:conceptual_flow}
\end{figure*}

In MIS, the subconscious assumptions typical of human visual perception can lead to erroneous interpretations of endoscopic and
radiological images, contributing to serious adverse events such as bile duct injuries that were rarer during traditional, open surgery procedures~\citep{mascagni2022artificial}. 
Therefore, understanding the surgical scene objectively can result in improved surgical outcomes. 

% \textcolor{orange}{Add some others, more recent, references}

Surgical scene semantic segmentation enables precise identification of anatomical structures and instruments, providing contextual information for comprehensive surgical scene understanding. Semantic segmentation partitions raw image data into structured and meaningful regions and thus enables further image analysis and quantification, which are critical for various applications, including anatomy research, disease diagnosis, treatment planning, and intra-operative decision support. Understanding the surgical scene via full scene segmentation is a prerequisite for many assistance functions in computer-assisted interventions (CAI), such as robotic assisted surgery and surgical workflow analysis. Surgical scene understanding can provide real-time surgical assistance in the operating room, thus improving the safety and efficacy of the procedure. 

% Such automated DL-based tools provide a good trade-off between speed and accuracy making it feasible to integrate such tools into a computer-aided surgical navigation systems and in robot-assisted surgery, among other application areas. Similarly to other domain areas, these advances have been possible due to the existence of large datasets in the area of surgical data science and access to vast computing resources. However, there are several challenges in real-world scenarios, and more in particular in the field of minimally invasive surgery, core among them are the robustness and generalizability to domain shifts \cite{koh2021wilds,philipp2022dynamic}, which make the translation of such solutions difficult. 

A vast majority of existing state-of-the-art (SOTA) computer vision models~\citep{yoon2022surgical,grammatikopoulou2024spatio} for surgical scene segmentation assume training and test data as independent and identically distributed (IID) i.e., data from same distribution. However, in real world settings, deploying models trained on data from one healthcare centre will suffer from performance degradation on data from other centres due to the domain shift caused by various factors such as patient demographics, instrument types, and vendor choice of the system to acquire the data. Fig.~\ref{fig:domain shift}(a) shows some of the intra-source and inter-domain variations separately. Typical endoscopic artifacts include the presence of smoke, specular reflections, illumination changes, and occlusions. Inter-domain variations comprise changes of aspect ratio between source and target data, photometric changes, changes in instrument types and vendors, and camera recording orientation. Fig.~\ref{fig:domain shift}(b) shows how color intensity distributions change across various healthcare centre datasets which inevitably can impact cross-centre model performance.  Furthermore, complications during surgical procedures, lighting conditions, surgeon expertise levels can also append to the domain shift resulting in even lower model performance. Work on domain-generalised surgical methods have been quite limited due to the unavailability of diverse multicentre annotated datasets. In this work, we aim to address these two important challenges, i.e., model generalisability and lack of multicentre annotated data. 

Efforts towards developing approaches for generalisable methods for surgical domain has recently gained attention with few studies~\citep{murali2023endoscapes,wagner2023comparative} published to tackle this problem. The advancements in this area of research have focused on two fronts: algorithmic and multicentre data collection level. On the algorithmic level, only the SOTA methods~\citep{wang2023sam,satyanaik2024optimizing} from the natural computer vision domain have been tested on a limited out-of-distribution (OOD) settings which fail to learn enough discriminative features from the training data limiting their capability to generalise well on the unseen distribution. On the multicentre data collection too, the progress has been limited, where in some cases, data has not been released publicly~\cite{satyanaik2024optimizing} to encourage further research and enable the research community to assess the generalisability of methods.

Therefore, the question we aim to answer is \textit{can we design ML models that can perform optimally on the IID samples but minimise performance degradation when OOD samples are encountered, i.e., generalisable to arbitrary unseen samples or distribution?} A generalisable model also alleviates the need of requiring access to multicentre data for training, which is otherwise unfeasible due to data protection and privacy restrictions.

To tackle the generalisability problem in semantic segmentation, several works have been carried out in the traditional computer vision (i.e., natural scenes). To this end, IBN-Net~\citep{pan2018two} combined batch and instance norm to learn domain-invariant features. Following this approach, several other methods have also used instance norm along with other feature learning blocks like feature whitening~\citep{choi2021robustnet}, category level learning~\cite{peng2022semantic}, style diversification~\citep{lee2022wildnet}, covariance alignment~\citep{ahn2024style}.    Although these methods have been successful in learning domain-invariant information, such techniques have not been widely explored in surgical computer vision settings. In this paper, we aim to bring the successes of existing DG approaches to the surgical domain to effectively learn domain discriminative features. Prior works~\citep{pan2018two,choi2021robustnet} have shown the effectiveness of using instance norm~\citep{ulyanov2016instance} in DG, however, IN alone is not adequate on two counts - 1) It merely standardises the features while also inevitably removing discriminant features~\citep{ahn2024style,jin2021style} as shown in Fig.~\ref{fig:conceptual_flow}(a). , and 2) it does not take into account the correlation between channels. The loss of discriminative information due to IN is further compounded in subsequent works using Instance whitening (Fig.~\ref{fig:conceptual_flow}(b).) and the combination of Instance norm and whitening (Fig.~\ref{fig:conceptual_flow}(c).). Furthermore, current DG approaches work in a multisource domain setting where multiple source domains are available for training. However, obtaining multiple labeled datasets is unfeasible for training especially in the surgical domain.

To address the above mentioned issues,  we propose a RobustSurg, a single domain generalization approach to tackle domain generalization in the surgical domain by learning from one source domain. RobustSurg consists of a  novel Domain-invariant feature encoder (DIFE) block that integrates style normalization and restitution (SNR) and instance selective whitening (ISW) block into the three convolution layers of ResNet-50 backbone to boost DG performance (Fig.~\ref{fig:conceptual_flow}(d).). SNR applies IN after the convolution layers to eliminate style variations and restores the discriminant features. Since feature covariance contains domain-specific information such as texture and color~\citep{gatys2016image,gatys2015texture}, the ISW block exploits the feature covariances from the original and augmented images through a whitening transformation (WT)~\citep{cho2019image} to selectively remove the style information and retain the structure and content. 

% \textcolor{orange}{I insist, here we need to conceptually need to show the main constributions of th paper}

Furthermore, a general lack of annotated multicentre test data to evaluate model generalization has been the single-most important bottleneck in the surgical DG domain. We overcome this limitation by taking a part of HeiChole Benchmark~\citep{wagner2023comparative} dataset comprising of data from two healthcare centres and labeling it with the scene segmentation labels, rigorously reviewed by an in-house clinical expert. We developed an annotation protocol based on the expert recommendations for cholecystectomy surgeries discussed in~\cite{mascagni2021surgical} and then a senior postgraduate student performed the pixel-wise labeling of the data followed by rigorous review conducted by an in-house clinical expert. We evaluated the effectiveness of our approach on both  in-domain test sets ($i.e.,$ IID) and on different modalities and different centre datasets ($i.e.,$ OOD). Specifically, we train the network on CholecSeg8k~\citep{hong2020cholecseg8k} dataset and test on our newly introduced multicentre test set. We will release both the implementation code and dataset upon acceptance of our work. To further validate model generalisability performance, we conducted model training and testing on two other dataset settings. One on multimodality EndoUDA dataset and another on cross-centre cataract surgery datasets. The EndoUDA~\citep{celik2021endouda} dataset is comprised of white-light imaging (WLI) and narrow-band imaging (NBI) for polyps and Barrett’s esophagus. WLI and NBI differ markedly in color distribution, contrast, and texture due to their distinct optical principles, with NBI emphasizing vascular and mucosal patterns through narrow wavelength illumination. Testing a WLI-trained model on NBI serves as a strong evaluation setup of cross-modality generalisability, assessing robustness to significant domain shifts common in real-world surgical deployments. Second, we conducted cross-hospital experiments on cataract surgery data and present results of a model trained on CaDIS dataset (recorded at the Brest university hospital) and tested on cataract-1K dataset (record at eye clinic of Klinikum Klagenfurt). We present evaluation results on different evaluation metrics such as mean Intersection over union (mIoU), precision, recall and pixel accuracy. We also provide per-class results to see how model behaves on instruments and anatomical regions. Further, we also perform different ablation experiments to see the effectiveness of various ISW and SNR blocks. Our major contributions are summarised below. 

% We address DG problem in surgical domain by style-content disentanglement through selective whitening (ISW block) and content restitution (SNR block)

\begin{itemize}
    \item We introduce a newly annotated cholecystectomy dataset with 8 classes from two surgical centres to rigorously test the domain generalization for surgical scene segmentation.
    \item We propose a novel domain-invariant feature encoder to address the loss of discriminant information from features in widely used instance normalization technique. 
    % by integrating instance selective whitening (ISW) and Style normalization and restitution (SNR) modules effectively, thereby learning domain invariant features.
    \item We validate our proposed approach for generalisability on three surgical  datasets. Our experiments demonstrate that our approach is superior to other state-of-the-art approaches. 
    
    % We propose a state-of-the-art deep learning  framework to address domain generalization problem for surgical scene segmentation. 
    % \item We integrate instance selective whitening (ISW) and Style normalization and restitution (SNR) modules to effectively learn domain invariant features. 
  
%    \item We evaluate the model effectiveness on various other medical domain datasets. 
   % \item We provide comprehensive ablation studies to validate the effectiveness of our approach. 
\end{itemize}  

%\subsection{Paper Organization}
The rest of the paper is organised as follows. Detailed related work on Domain generalization in natural and surgical domain and discussion about lack of diverse datasets is discussed in Section~\ref{sec:Related Work}. Details of our new multicentre dataset and method are explained in Section~\ref{sec:materials and method}. Section~\ref{sec:implementation_details} presents the implementation details,  experimental setup and the description of evaluation metrics used. Results are comprehensively presented and discussed in Section~\ref{sec:results-discussion}. Finally, conclusion and future work is presented in Section~\ref{sec:conclusion}.

%Furthermore, to enable future robotic assisted surgery systems, they will have to understand the surgical scene, therefore, surgical scene understanding is the prerequisite for the next generation of surgical robotics.

% \textcolor{blue}{I think is important to clearly show how computer vision is used in clinical setting such as MIS or laparoscopy and show particular challenges so it clearly to the reader, and briefly show what are the expected outcomes (i.e.,s segmentation) in a conceptual way... this about a graphical abstract here...}

\begin{figure*}[t]
    \centering
\includegraphics[width=\linewidth]{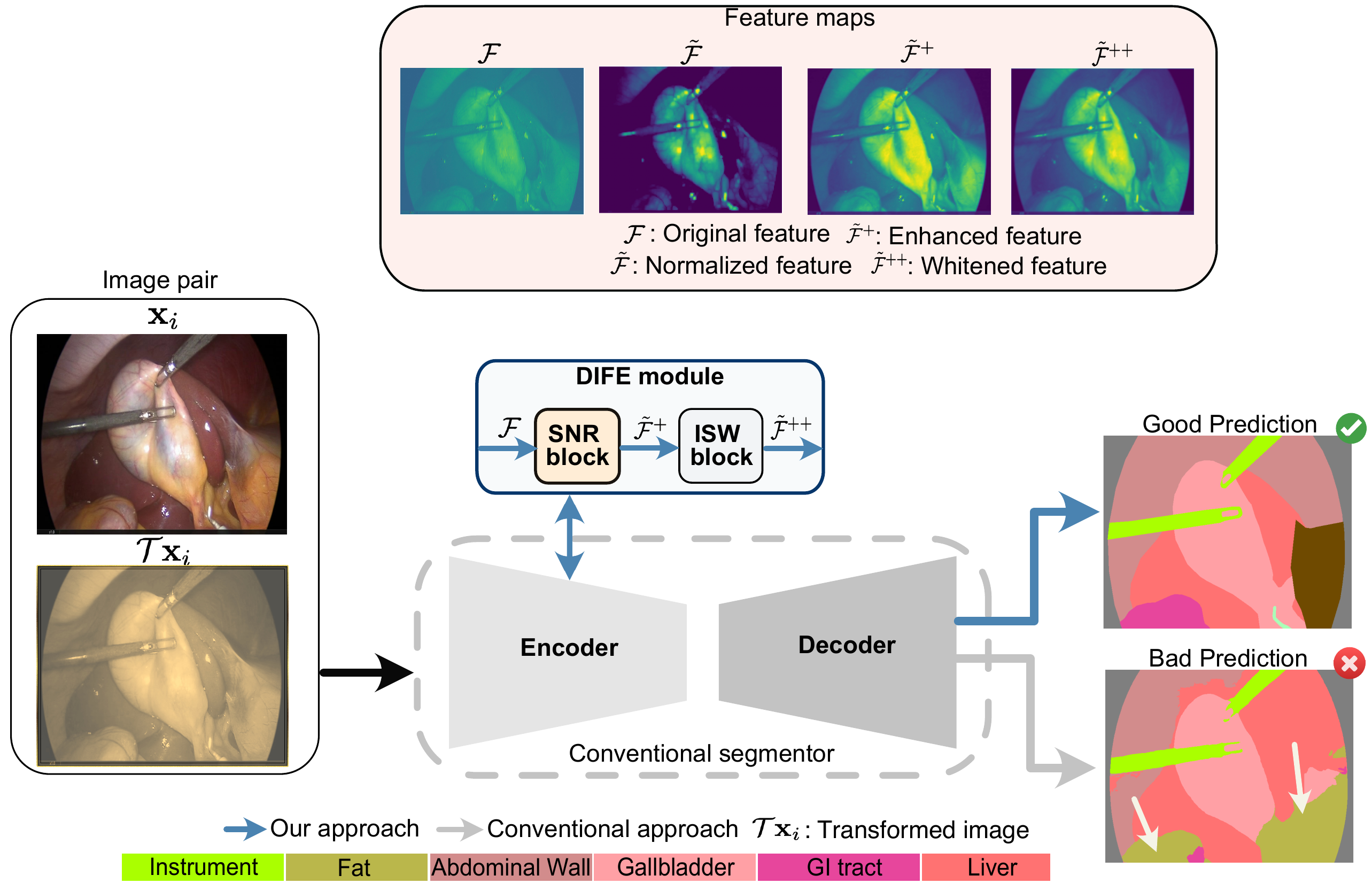}
    \caption{Block diagram of the RobustSurg method for generalisable surgical scene segmentation.  The encoder takes two images, i.e., raw image and transformed image. Conventional segmentors such as DeepLabv3+~\citep{chen2018encoder} underperform on unseen images as shown in the prediction. In our approach, we introduce domain-invariant feature encoder (DIFE) module containing two sub-blocks. SNR block~\citep{jin2021style} normalises and recovers the lost features while ISW block selectively suppresses the style information. }
    \label{fig-framework}
\end{figure*}

\section{Related Work}
\label{sec:Related Work}

%\subsection{Related Concepts - look better word}

Learning domain-agnostic models from limited source domain data is an important area of investigation in several fields, including computer vision at large and medical imaging in particular. In this section, we revisit some of the related works in the field domain generalization in natural and medical image scenarios. 

%\textcolor{orange}{Figure 2 is barely used to discuss any points in the section, you need to make better use of the conceptual elements}

\noindent \textbf{Domain Generalization. } Domain generalization aims to extract domain-invariant features from source domain(s) to learn a generalisable representation better-suited for unseen target domains. DG methods can be broadly categorised into three different groups: (i) Representation learning techniques that aim to extract domain-invariant knowledge from the source domain(s) that seamlessly generalise to any arbitrary target domain \citep{blanchard2011generalizing,nguyen2021domain}. (ii) Data augmentation methods learn to diversify the source domains to have more representative training data~\citep{gong2019dlow,zhou2020learning,zhou2021domain} (iii) Learning-strategy approaches may achieve generalization through meta-learning~\citep{wei2021metaalign}, self-supervised learning~\citep{carlucci2019domain}, causality learning~\citep{sun2021recovering} or optimization procedures~\citep{cha2021swad} aimed at finding a flat minima.  In this work, we approach the DG problem using the representation learning approach where we aim to separate the style and content information learn the generalisable representation. 

\noindent \textbf{Style-Content Disentanglement.} S-C disentanglement aims to partition the aggregated latent variables into two separate components, denoted as style and content. While the concepts of style and content have their origins in the style transfer literature~\citep{mathieu2016disentangling,szabo2017challenges}, recent studies have further advanced the concept by incorporating the approaches like causal inference~\citep{peters2017elements}, independent component analysis~\citep{gresele2021independent}, and feature covariance~\citep{choi2021robustnet}. CDDSA~\citep{gu2023cddsa} proposed separate anatomy encoder and modality encoder to disentangle style-content features and used the decoder to reconstruct the final prediction. Style-semantic contrastive loss was proposed to disentangle the source and target stylised features~\citep{guo2023style}. Style-content disentanglement was tackled by adding covariance alignment modules in the encoder and contrastive learning in the decoder for DGSS~\citep{ahn2024style}. In a multisource DG settings~\citep{lee2022wildnet} where multiple source domains are employed as training data, disentanglement of domain-invariant features have shown great potential in training robust deep learning models by leveraging shared knowledge across domains. On the contrary, disentanglement has been rarely studied in single DG (sDG) setting where just one source domain is available for training. 

Previous literature in sDG \citep{gatys2016image,choi2021robustnet} reveal that feature correlation (i.e., a covariance matrix) stores the domain-specific styles of images. Specifically, the whitening transformation has shown success in removing feature correlation and domain-specific style information in image translation \citep{cho2019image}, style transfer \citep{li2017universal}, domain adaptation \citep{roy2019unsupervised} and generalization \citep{choi2021robustnet}, thus improving model generalisability.   
It has also been shown that instance normalization (IN) in shallow layers of CNNs boost model's ability to learn discriminant features~\citep{pan2018two,nam2018batch}.  However, explicit use of IN can result in loss of discriminant information, since it is task-agnostic. . 
 Building upon previous work on feature disentanglement \citep{jin2021style} and feature whitening \citep{choi2021robustnet}, we aim to disentangle style-content information in a sDG setting for surgical scene segmentation using the combination of feature covariance and style restitution to improve DG performance.

% \textcolor{orange}{This has to be in separate sub-section:
\textbf{Domain generalization in surgical domain.} Domain generalization has largely been an under-explored topic in surgical domain. There are only a few studies conducted on learning domain-agnostic models for surgical instrument localization~\citep{philipp2022dynamic,philipp2021localizing}, scene understanding~\citep{seenivasan2022biomimetic}, object-centric learning~\citep{satyanaik2024optimizing}, and endoscopic image segmentation~\citep{guo2024infproto}. Despite the initial efforts, DG in the surgical domain remains methodologically underexplored, with limited efforts targeting domain-agnostic learning for critical tasks such as full scene segmentation. In terms of datasets, the Heidelberg colorectal dataset which was successfully used in the Robust-MIS~\citep{ross2020robust} challenge was the first comprehensive multicentre dataset to assess model generalisability for instrument detection and segmentation under challenging surgical images. ~\cite{bar2020impact} investigated the generalisability of surgical workflow recognition task on a multicentre surgical dataset showing that training models on diverse datasets can potentially enable translating AI research into the clinical practice. Although these datasets are based on clinical patient videos for their annotations, their applicability is limited, as they do not adequately capture the variability of clinical data across multiple centres and focus on only on specific features of the surgical images (e.g surgical instruments). To address these limitations, the Heichole multicentre dataset~\citep{wagner2023comparative} was initially introduced for surgical workflow analysis, and skill assessment and later used in EndoVis 2021 challenge for surgical scene segmentation. However, the scene segmentation annotations were not publicly released. 

The lack of model generalization works in surgical domain can be partly attributed to the scarcity of labeled multicentre datasets for model evaluation.  One line of work to address data scarcity problem in surgical domain has proposed diffusion models to generate diverse annotated datasets~\citep{venkatesh2025data}. However, recent research~\citep{akbar2025beware} has shown that instead of learning the structure of data, diffusion models are more prone to learn the training data itself and do not produce diverse data. We derive our motivation from this important observation and introduce an annotated multicentre dataset by taking publicly available HeiChole multicentre videos~\citep{wagner2023comparative} and annotating with semantic segmentation labels for testing model generalisability in terms of surgical scene segmentation.

% Guo et.al \cite{guo2024infproto} proposed an agnostic feature learning approach which aims to capture domain-invariant features by minimizing domain-specific biases. 

% Pan et al. used combination of IN and Batc|h Normalization (BN) together (half of channels use IN while the other half of channels use BN) in the same layer to preserve some discrimination. Nam et al. [41] determine the use of BN and IN (at dataset-level) for each channel based on learned gate parameters. However, It lacks the adaptivity to instances. Besides, the selection of IN or BN for a channel is hard (0 or 1) rather than soft.

% \textcolor{cyan}{it could be interesting to briefly discuss some previous datasets on previous works such as robust mis and areas of opportunity such as your data}

% \textcolor{cyan}{Try to focus on what is in the state of the art and be super precise about the lack of datasets with multi-center data!!!}

% \textcolor{cyan}{Overview, you have just discussed the state of the art! In the summary of the state of the art you have to be very clear of what is missing... }

%\subsection*{Limitations} 
\textbf{Limitations.} While Domain generalization has witnessed growing interest in the natural image domain, its applications to the surgical domain still remain underdeveloped in terms of methodological rigor and lack of data diversity.  Existing efforts in surgical domain have largely focused on a few tasks such as instrument localization, or segmentation, but they lack robust frameworks for generalization across diverse surgical environments. In this work, we aim to adapt Style-content disentanglement, one of the effective approaches in natural DG to the surgical domain and aim to alleviate the loss of information due to instance normalization with feature restitution.

A great obstacle in developing generalisable methods for laparoscopic surgery is the lack of diverse multicentre datasets. To address this gap and to encourage further research, we release HeiCholeSeg, a pixel-wise annotated dataset from two different surgical centres for generalisability evaluation of deep learning models. To summarise, domain-agnostic approaches tailored to the domain shifts in surgical data, alongwith multicentre annotated datasets are needed to further advance the research in surgery.

\section{Materials and Method} \label{sec:materials and method}
%

% \textcolor{cyan}{I think that this section requires:}

% \textcolor{cyan}{1. A general overview section, in which you present the main components of the proposed method, the specific tasks (segmentation) and the challenges, then your introduce briefly the dataset you created...  }

% \textcolor{cyan}{2. A preliminaries section. In which you discuss the main concepts behind DG, what are relevant and irrelevant features and how your are building upon previous works}

% \textcolor{cyan}{3. Then you can describe the each of the components of the architecture and the details: what was is the basics and which are the contributions}

In this section, we first delve into the details of the novel HeiCholeSeg, multicentre annotated dataset, and the annotation and validation process. The next part of the section presents the RobustSurg method. 

\subsection{Novel multicentre annotated surgical dataset} \label{heicholeseg details}

\begin{figure*}[t!]
    \centering
\includegraphics[width=0.6\linewidth]{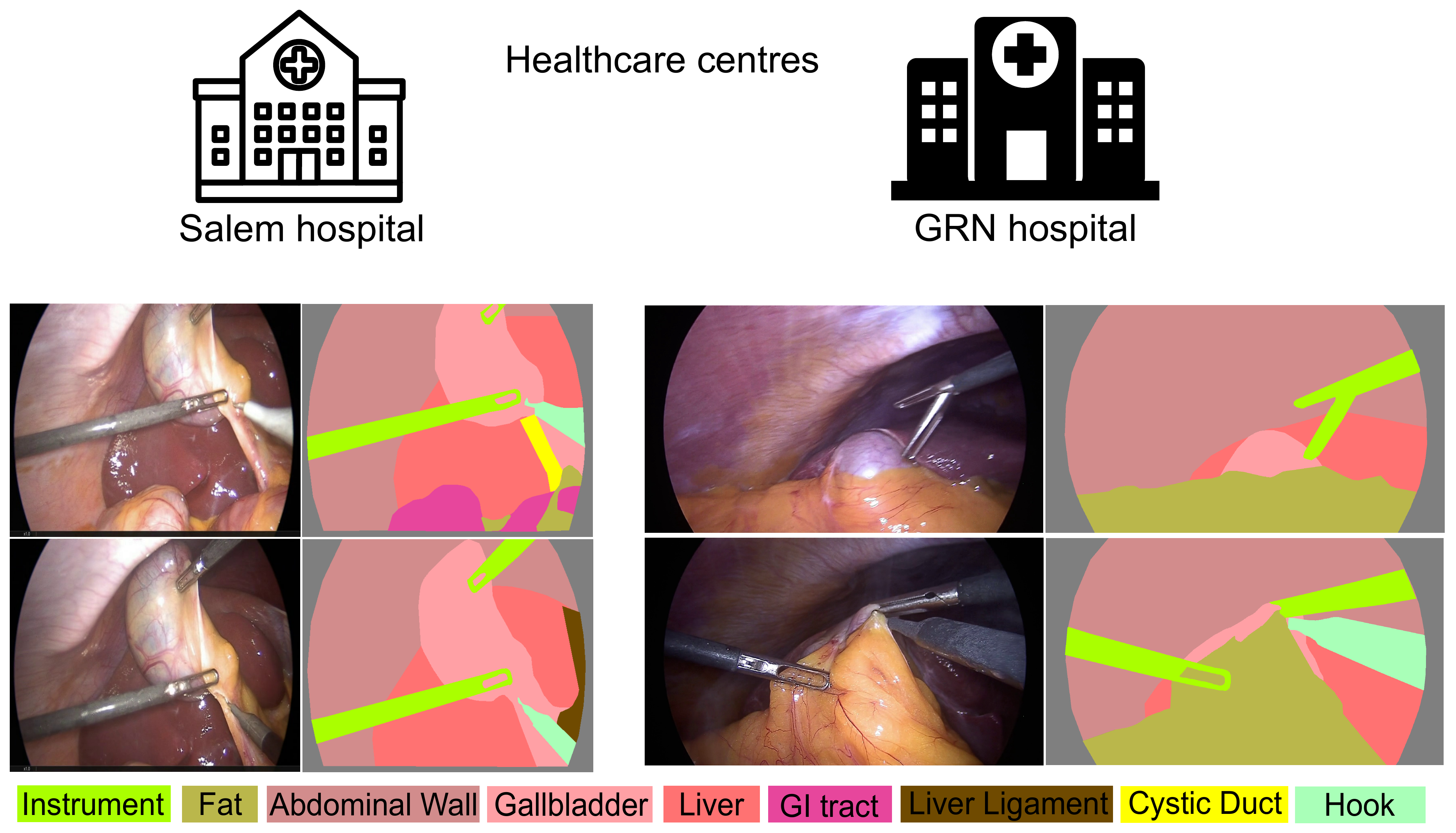}
    \caption{Example from the HeiCholeSeg multicentre dataset. Each image is shown alongwith its corresponding semantic segmentation mask. Images depict varying photometric properties, and aspect ratios across the surgical centres.  }
    \label{fig:examples_frames}
\end{figure*}

We introduce HeiCholeSeg, a multicentre pixel-wise annotated dataset to address the limitation of lack of diverse, representative and annotated datasets in the surgical domain. The HeiCholeSeg dataset is taken from the public HeiChole Benchmark~\citep{wagner2023comparative} which consists 24 public videos of laparoscopic cholecystectomies performed at three German hospitals. The videos at Heidelberg University Hospital were recorded with a laparoscopic 2D camera (Karl Storz SE \& Co KG, Tuttlingen Germany), and at Salem and GRN-hospital Sinsheim hospitals, they were recorded with the laparoscopic 2D camera ENDOCAM Logic HD (Richard Wolf GmbH, Knittlingen, Germany). Originally, the dataset was frame-wise annotated with surgical phases, actions, instrument presence and skills but did not include segmentation masks. 

% \begin{table*}[htbp]
% \centering
% \caption{Overview of data acquisition settings across different medical centers}
% \label{tab:data_sources}
% \begin{tabular}{p{3cm}p{5.5cm}p{2.5cm}p{1.5cm}p{2.5cm}}
% \toprule
% \textbf{Data Center} & \textbf{Laparoscope} & \textbf{Resolution} & \textbf{FPS} & \textbf{No. of images} \\
% \hline
% Heidelberg University Hospital &
% 2D camera (Karl Storz SE \& Co KG, Tuttlingen, Germany) with 30$^\circ$ optics &
% 960 × 540 &
% 25 &
% 100
% \\

% \hline
% Salem, Heidelberg &
% 2D camera ENDOCAM Logic HD (Richard Wolf GmbH, Knittlingen, Germany) with 30$^\circ$ optics &
% 720 × 576 &
% 50 &
% 150
% \\
% \hline
% GRN-hospital Sinsheim &
% 2D camera ENDOCAM Logic HD (Richard Wolf GmbH, Knittlingen, Germany) with 30$^\circ$ optics &
% 1920 × 1080 &
% 50 &
% 150
% \\
% \hline
% Strasbourg &  &   480 × 854 & 25 & 8080 \\
% \hline
% \end{tabular}
% \end{table*}

A subset of frames was selected for annotation, since annotating all 24 videos would have resulted in more than 10 million images and an impractical labeling effort. To this end, two full surgery videos (each having duration of nearly an hour and each from Salem and GRN hospitals) were selected for the HeiCholeSeg data based on the annotation capacity. Details regarding the dataset characteristics can be found in Table~\ref{tab:data_sources} and some example frames are provided in Fig.~\ref{fig:examples_frames}. To ensure consistency and surgical phase alignment between training and test datasets, an expert clinician manually selected the relevant temporal segments of each video. These segments correspond specifically to the cholecystectomy phase of the surgery. This step was crucial for creating a consistent benchmark across institutions and to match the training dataset phases. From the selected video segments, frames were extracted at a fixed sampling rate of 25 frames per second (fps) (with a custom python script). This rate was chosen to balance annotation burden and frame data variability. A fourth year postgraduate student having 4 years of experience in artificial intelligence in surgical image analysis guided by a medical expert with over 6 years of experience proceeded  with the labeling task. Annotation was performed using Wacom tablets to facilitate precise and efficient pixel-level delineation of semantic classes on Labelbox~\citep{labelbox}, an open-source collaborative web-based tool.

The annotator was required to annotate all the semantic classes including anatomical structures and surgical instruments in the frames. The annotator conducted the initial round of annotations on selected frames from cholecystectomy videos from each centre. The medical expert reviewed the annotated data, providing detailed feedback and comments on instances of potential mislabeling, ambiguity, or anatomical/surgical inaccuracy. Based on the expert’s feedback, the annotations were revised to correct the class delineations and anatomy class assignments. The revised annotations underwent a second round of review by the medical expert who provided feedback on any remaining inconsistencies. The flagged instances were returned to the annotator for a second re-annotation. Following this re-annotation, the medical expert performed final validation of the annotations. Although some delineation errors are inevitable given the inherent complexity of segmentation tasks, our iterative, multistage annotation protocol substantially reduced these errors, resulting in high-quality and reliable annotations.

Detailed annotation instructions as well as class definitions were determined and communicated in the annotation protocol. Some of the general rules for the annotation task are as follows:

\begin{itemize}
    \item The cystic duct and the cystic artery are two tubular structures which are connected to the gallbladder. 
    \item Gallbladder is a pear-shaped organ located beneath the liver,  appears as grey-blue sac attached to the liver surface.
    \item Liver is a large reddish-brown organ occupying the right upper abdomen, partially visible above the gallbladder.
    \item  Segment the region visible through the fenestrated jaws of instruments such as grasper. 
\end{itemize}

\begin{table*}[htbp]
\centering
\caption{Overview of data acquisition settings across different medical centres. Note that CholecSeg8k data was recorded at Strasbourg centre.  }
\label{tab:data_sources}
\begin{tabular}{p{0.5cm}p{3cm}p{5.5cm}p{2.5cm}p{0.5cm}p{2.0cm}}
\toprule
 & Data Centre & Laparoscope & Resolution & FPS & \# of images \\

\midrule
\multirow{3}{*}{\raisebox{-1.5ex}{\rotatebox{90}{HeiCholeSeg}}}

% & Heidelberg University Hospital &
% 2D camera (Karl Storz SE \& Co KG, Tuttlingen, Germany) with 30$^\circ$ optics &
% 960 × 540 & 25 & 100 \\

& Salem hospital &
2D camera ENDOCAM Logic HD (Richard Wolf GmbH, Knittlingen, Germany) with 30$^\circ$ optics &
720 × 576 & 50 & 150 \\

& GRN-hospital Sinsheim &
2D camera ENDOCAM Logic HD (Richard Wolf GmbH, Knittlingen, Germany) with 30$^\circ$ optics &
1920 × 1080 & 50 & 150 \\

\hline
%\multirow{1}{*}{\rotatebox{90}{CholecSeg8k}} 
& Strasbourg centre &  & 480 × 854 & 25 & 8080 \\

%& \multicolumn{5}{c}{\textbf{Cataract surgery}} \\
\bottomrule
\end{tabular}
\end{table*}

\subsection{Method}

The main goal of this work is to train a semantic segmentation model on source domain (surgical endoscopic data from one distribution or a healthcare centre) which generalises well on target domain (unseen distribution or different healthcare centre), performing better than the baseline conventional model i.e., DeepLabv3+~\citep{chen2018encoder}. Conventional segmentation models lack the ability to learn sufficiently discriminative features from the training data hence their performance severely degrades when tested on OOD data. Therefore, we tackle this problem by introducing a plug-and-play domain-invariant feature encoder (DIFE) block which which can be plugged to any backbone network such as ResNet50 of DeepLabv3+. Fig.~\ref{fig-framework}  shows the overall proposed framework of RobustSurg and its comparison to conventional models for semantic segmentation. The DIFE module in our proposed method consists of two blocks adapted from the state of the art to the surgical domain, Style normalization and Restitution (SNR)~\citep{jin2021style},  and instance selective whitening (ISW)~\citep{choi2021robustnet}. The DIFE module works by performing three operations on each feature map passed from an encoder layer. The first task being style normalization through instance norm, the second task is the restitution operation to restore the discriminant information lost during instance norm, and then finally the whitening operation to selectively whiten the domain-dependent styles. We also tackle the lack of multicentre annotated datasets by introducing HeiCholeSeg, a two-centre annotated dataset to evaluate the generalisability of models.

% The overall proposed framework is described in Fig.~\ref{framework}. First we extract features from ResNet50 backbone which are then passed to the SRW module containing SNR and ISW blocks. In SNR block instance normalization (IN) is applied to standardize the features, while SNR module recovers the lost domain-invariant information due to IN. Below we detail both SNR and ISW blocks in detail. 
% therefore, in this work we exploit this method for our experiments.
%
% \begin{figure*}[t]
%     \centering
% \includegraphics[width=\linewidth]{img/overview.pdf}
%     \caption{Block diagram of the RobustSurg method for generalizable surgical scene segmentation. \textbf{A.} depicts the overall flow of the method trained on two datasets. The encoder takes two images, i.e., raw image transformed image, Initially, and feeds the features to the SRW block.  \textbf{B.} depicts, SNR block \cite{jin2021style} selectively suppresses the style information followed by up-sampling network,  while WT applied selectively results in improved domain-specific and domain-invariant features.  }
%     \label{framework}
% \end{figure*}
%

% selectively removing the style information and retaining the content. We use the feature covariance matrix from a clean image and a photometric transformed image to achieve this task. 
%
\subsection{Style Normalization and Restitution (SNR block)}

% \textcolor{cyan}{From the general overview, it has to be clear why we want to use the SNR module... So far we haven't really given the reader a clear uderstanding... I thiunk this has to be clear at least conceptually... Just throwing the description of the module does not help.....It is necessary to provide a clear and logical flow of ideas}
\begin{figure*}[t!]
    \centering
\includegraphics[width=0.8\linewidth]{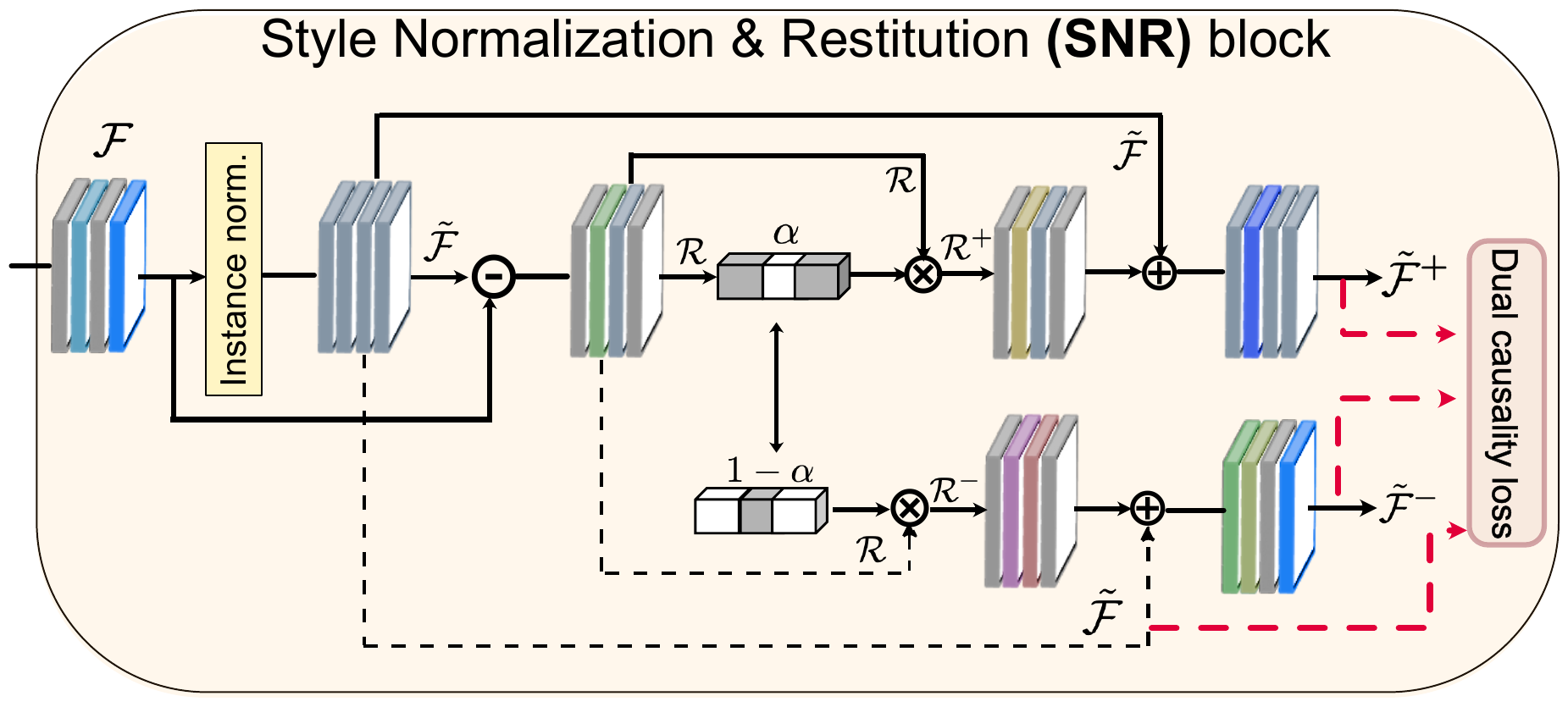}
    \caption{Features from intermediate backbone layers are fed to the SNR block. SNR applies instance normalization (IN) on input features, followed by channel attention to restore lost information due to IN. Dual causality loss encourages the disentanglement  between useful and contaminated features.}
    \label{fig:snr}
\end{figure*}

Several previous works \citep{pan2018two,li2017universal} have shown that instance normalization is less sensitive to style variations. Most of the earlier DG methods have used IN alongwith feature statistics (e.g. mean, variance, covariance) to extract the style information aiming to suppress domain-dependent styles and preserve the content for better generalisability. The problem with IN is that it lacks task awareness and results in the loss of discriminant information necessary for better performance \cite{li2017universal}. To tackle this problem, we incorporate the Style Normalization and Restitution (SNR) block which aims to restore the lost information due to IN by applying channel attention. Channel attention aims to decouple the task-relevant and non-relevant features on channel dimensions to be consistent with the IN which also works on the feature channels. In the SNR block, features are first instance normalised, followed by a feature restoration process where the task-relevant and non-relevant features are distilled through channel attention. 

%$\widetilde{\mathcal{F}}$
%The SNR module enhances the generalization capability of our model while preserving the discriminative power of the segmentation network for achieving an effective DG.
Fig. \ref{fig:snr} shows the overall flowchart of the SNR block originally proposed in~\citep{jin2021style}. The SNR layer receives an input feature map  $\mathcal{F}\in \mathbb{R}^{N \times C \times H \times W}$ and outputs an enhanced map $\widetilde{\mathcal{F}}^+\in \mathbb{R}^{N \times C \times H \times W}$. Instance normalization is first applied to eliminate style discrepancies in the features maps. It is followed by a restitution step where the task-specific features from the residual are restored (i.e., the difference between style normalised $\widetilde{\mathcal{F}}$ and original feature $\mathcal{F}$ maps). This restoration is carried out by masking  $\mathcal{R}$ with the channel attention vector $\boldsymbol{\mathcal{\alpha}} = [\mathcal{\alpha}_1,\mathcal{\alpha}_2,..\mathcal{\alpha}_C] \in \mathbb{R}^C$. This masking operation is implemented using spatial global average pooling and two fully-connected (FC) layers. Applying channel attention, we get,  
\begin{equation}
    \mathcal{R}^+(:,:,:,i) = \mathcal{\alpha}_i \mathcal{R}(:,:,:,i)
\end{equation}
\begin{equation}
    \mathcal{R}^-(:,:,:,i) = (1-\mathcal{\alpha}_i) \mathcal{R}(:,:,:,i)
\end{equation}
where $\mathcal{R}(:,:,:,i)$ represents the $i^{th}$ channel of the feature map $\mathcal{R}$.  On the other side of Eqs., (1) and (2), $\mathcal{R}^+$ and $\mathcal{R}^-$ denote the task-relevant and irrelevant features, respectively. Enhanced ($\widetilde{\mathcal{F}}^+$) and corrupted feature maps $(\widetilde{\mathcal{F}}^-$) are then obtained by adding $\mathcal{R}^+$ and $\mathcal{R}^-$ to normalised features $\widetilde{\mathcal{F}}$, respectively. We integrate dual causality loss $\mathcal{L}_{dc}$ function which is basically an entropy minimization loss constraint to enhance the disentanglement of task relevant and non-relevant features. 

%\textcolor{orange}{why? is this something new? how does improve the orignal forumations of the SNR block? why is not listed in the contributions?}

With the addition of $\mathcal{R}^+$ to $\widetilde{\mathcal{F}}$,  enhanced feature map $\widetilde{\mathcal{F}}^+$ becomes more discriminative resulting in smaller entropy (pixel randomness or uncertainty), in contrast to the corrupted feature maps which have larger entropy. Dual Causality loss aims to enhance the separation between enhanced and corrupted features. In $\mathcal{L}_{dc}$, first we compute pixel-wise entropy with the function $\mathcal{E}(.) = -p(.)\log{p}$, where p is denotes softmax probability. The $\mathcal{L}_{dc}$ consists of $\mathcal{L}_{SNR}^+$ and $\mathcal{L}_{SNR}^-$ i.e.,  $\mathcal{L}_{SNR}^+$ + $\mathcal{L}_{SNR}^-$,   

% \begin{equation}
%     \mathcal{L}_{SNR}^+ = Marginloss(\frac{1}{w\times h} \sum_{j=1}^{h} \sum_{k=1}^{w} \mathcal{E} (\phi(\widetilde{\mathcal{F}}^+(j,k,:))) - \frac{1}{w\times h} \sum_{j=1}^{h} \sum_{k=1}^{w} \mathcal{E} (\phi(\widetilde{\mathcal{F}}(j,k,:))))
% \end{equation}
% \begin{equation}
%     \mathcal{L}_{SNR}^- = Marginloss(\frac{1}{w\times h} \sum_{j=1}^{h} \sum_{k=1}^{w} \mathcal{E} (\phi(\widetilde{\mathcal{F}}(j,k,:))) - \frac{1}{w\times h} \sum_{j=1}^{h} \sum_{k=1}^{w} \mathcal{E} (\phi(\widetilde{\mathcal{F}}^-(j,k,:))))
% \end{equation}

\begin{equation}
\begin{split}
    \mathcal{L}_{SNR}^+ = \text{Marginloss}\Bigg(&\frac{1}{w\times h} \sum_{j=1}^{h} \sum_{k=1}^{w} \\
    &\mathcal{E} \left( \phi(\widetilde{\mathcal{F}}^+(j,k,:)) - \phi(\widetilde{\mathcal{F}}(j,k,:)) \right) \Bigg)
\end{split}
\end{equation}

\begin{equation}
\begin{split}
    \mathcal{L}_{SNR}^- = \text{Marginloss}\Bigg(&\frac{1}{w\times h} \sum_{j=1}^{h} \sum_{k=1}^{w} \\
    &\mathcal{E} \left( \phi(\widetilde{\mathcal{F}}(j,k,:)) - \phi(\widetilde{\mathcal{F}}^-(j,k,:)) \right) \Bigg)
\end{split}
\end{equation}
% \begin{equation}
%     \mathcal{L}_{SNR}^- = \text{Marginloss}(\frac{1}{w\times h} \sum_{j=1}^{h} \sum_{k=1}^{w} \mathcal{E} (\phi(\widetilde{\mathcal{F}}(j,k,:)) - \phi(\widetilde{\mathcal{F}}^-(j,k,:))))
% \end{equation}
where $\phi $ denotes softmax function and w, h represent height and width of feature vector containing probability values. On the other hand, Marginloss is $\ln{(1+\textit{exp}(.))}$ represents a monotonically increasing function whose aim is to serve as smoother optimiser to avoid negative loss values.

\subsection{Instance Selective Whitening (ISW block)}\label{isw}

% \textcolor{cyan}{Same here, it has to be clear what the SNR and the ISW modules are doing individually and how they improve the overall performance when combined. Why nobody has combined these ideas before?}

It has been shown that the whitening transformation (WT) can remove domain-specific information and improve the overall DG performance~\citep{cho2019image,li2017universal,pan2019switchable}. For a feature map $\mathcal{F}\in \mathbb{R}^{N \times C \times H \times W}$, WT is a linear transformation which standardises features by de-correlating the channels. It performs this de-correlation by making the feature covariance matrix ($\theta_s$) close to an identity matrix. 
The conventional way of computing WT through eigen value decomposition is highly computationally expensive. Thus  an alternative implementation proposed in GDWCT~\citep{cho2019image} is implemented that computes the deep whitening transformation (DWT):
\begin{equation}
    \mathcal{L}_{\mathrm{DWT}} = \mathbb{E} [\| \theta_\mu - \mathrm{I} \|_1]
\end{equation}
where $\mathbb{E}$ denotes the arithmetic mean.

\begin{figure*}[t!]
    \centering
\includegraphics[width=0.8\linewidth]{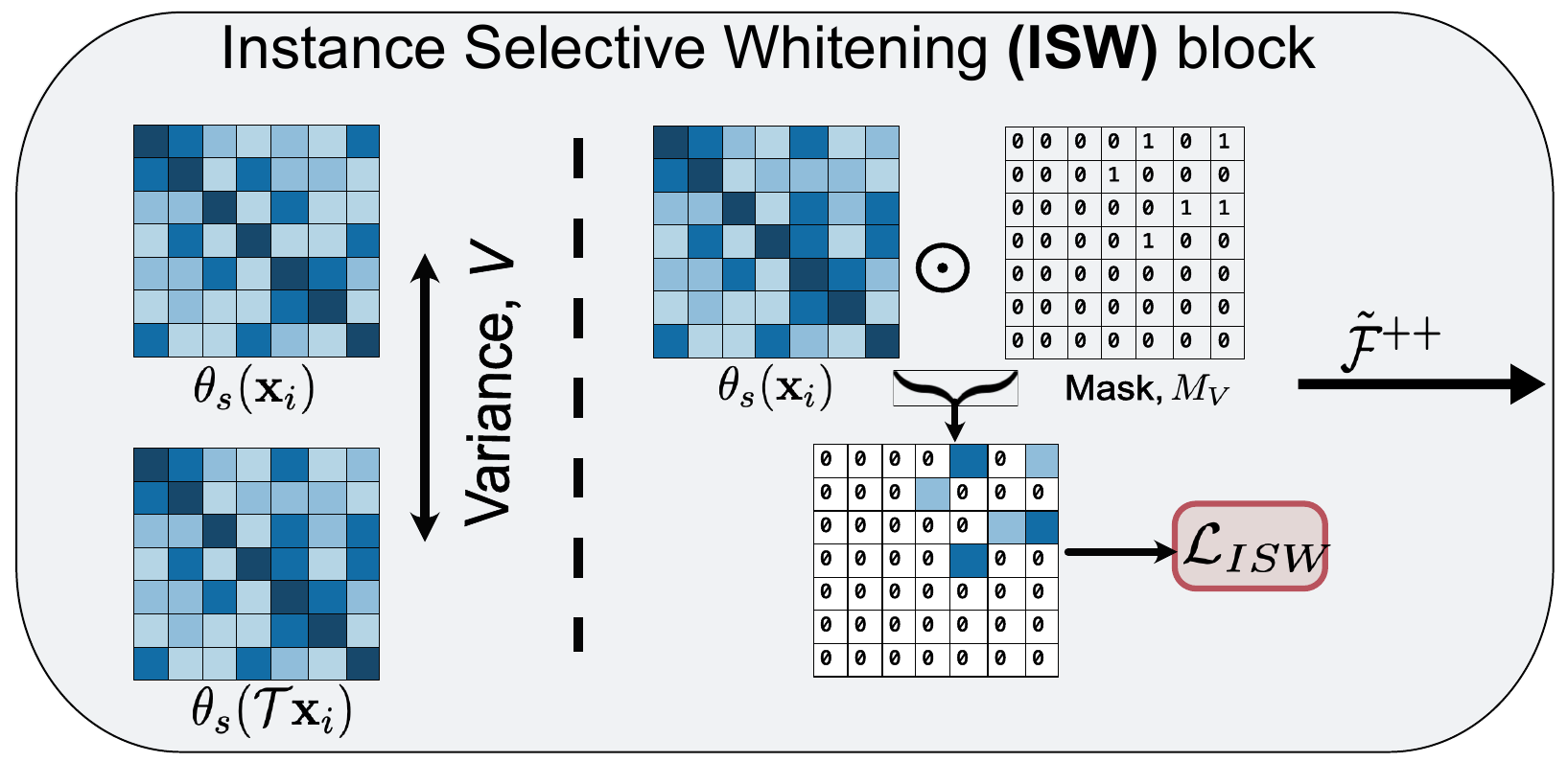}
    \caption{Instance selective whitening takes enhanced features from the SNR to selectively suppress the style-encoded information through instance selective whitening loss.}
    \label{fig:isw}
\end{figure*}

The major limitation of WT is that it can distort the object boundaries~\citep{li2017universal} or reduce feature discrimination~\citep{pan2019switchable} because $\theta_s$ contains both style and content information. Therefore, in our framework we introduce the ISW block to selectively remove the style while retaining the structural information (see Figure~\ref{fig:isw}).

Unlike prior approach~\citep{choi2021robustnet} on DG, where feature whitening is applied directly on the instance normalised features, we pass the distilled features from the SNR block to the ISW block. The ISW block takes the enhanced features $\widetilde{\mathcal{F}}^+$ of the original and transformed images from the SNR block. The network is initially trained for $n$ epochs ($n$ was empirically set to 5) to obtain stable covariance matrices. Afterwards, the covariance $\theta_s$ for both feature maps, as well as their variance V are computed as follows: 
\begin{equation}
    \theta_s = \frac{1}{h \times w} (\widetilde{\mathcal{F}}^+)(\widetilde{\mathcal{F}}^{+})^T
\end{equation}

\begin{equation}
    V = \frac{1}{N} \sum_{i=1}^N \frac{1}{2} ((\theta_s (x_i) - \mu_\theta)^2 + (\theta_s(\mathcal{T}x_i) - \mu_\theta)^2 )
\end{equation} 
\noindent where $\mu_\theta$ denotes the mean of both covariance matrices. It is assumed that high variance values in V indicate the presence of style information which must be suppressed. Therefore, a mechanism to disentangle and separate such values  was implemented using k-means clustering. This results in two distinct groups, $G_{high}$ (containing domain style) and $G_{low}$ (containing content information). Based on this clustered V, we compute the mask $M_v$ and consequently $\mathcal{L}_{ISW}$ as follows, 
\begin{equation}
    \mathcal{L}_{ISW} = \mathbb{E} [| \theta_s \textstyle\bigodot \mathcal{M}_v |]
\end{equation}
The overall cost function for our proposed approach is given by, 

\begin{equation}
    \mathcal{L}_{total} = \mathcal{L}_{task}+  \sum_i^L \lambda_1 \mathcal{L}_{ISW}^i + \lambda_2 \mathcal{L}_{dc}^i 
\end{equation}
with $\lambda_1$ and $\lambda_2$ denote the weight of the ISW and DC losses, respectively, L represents the number of layers and $\mathcal{L}_{task}$ is cross-entropy loss for semantic segmentation.

\begin{table*}[t!]
% \footnotesize
\caption{Table showing mean Intersection over Union (mIoU), Dice, Precision, Recall and accuracy scores for the SOTA and our proposed method on CholecSeg8K and HeiCholeSeg datasets. The best and second best results are \textbf{bolded} and \underline{underlined}, respectively.}\label{tab1:results}
\begin{tabular}{lccccc}
% \textbf{EndoUDA (polyp)}&\multicolumn{2}{c}{\textbf{Source Modality}}& \multicolumn{1}{c}{\textbf{Target Modality}}\\
\toprule
In-domain distribution&\multicolumn{5}{c}{CholecSeg8K Test set}\\
\hline
Method & mean IoU & mean DSC &  Precision & Recall & Pixel Accuracy \\
\hline
Baseline~\cite{chen2018encoder} & $87.67\pm0.624$ & $92.45\pm0.432$ &$\underline{94.57} \pm \underline{0.380}$ & $91.66 \pm 0.450$ & $97.78 \pm 0.190$  \\

IBN-Net~\cite{pan2018two}& $86.82\pm0.433$ & $91.63\pm0.320$&$93.26 \pm 0.290$ & $92.00 \pm 0.330$ & $97.76 \pm 0.160$  \\

RobustNet~\cite{choi2021robustnet}   & $87.67 \pm0.566$ & $92.52\pm0.411$ & $93.25 \pm 0.370$ & $92.31 \pm 0.390$ & $97.85 \pm 0.170$ \\

SNR~\cite{jin2021style} & $\underline{88.30} \pm\underline{0.423}$ & $91.89\pm0.450$ &  $93.14 \pm 0.410$ & $\underline{93.84} \pm \underline{0.420}$ & $\underline{97.90} \pm \underline{0.170}$  \\

SAN-SAW~\cite{peng2022semantic} & $82.54 \pm0.462$ & $89.88\pm0.476$ & $91.69 \pm 0.450$ & $88.64 \pm 0.480$ & $96.16 \pm 0.210$ \\

SHADE~\cite{zhao2022style} & $86.78 \pm0.503$ &  $\underline{93.45} \pm\underline{0.357}$ & $91.58 \pm 0.320$ & $92.47 \pm 0.391$ & $97.60 \pm 0.178$ \\

SAMed~\cite{SAMed} & $86.00 \pm0.550$ &  $91.24\pm0.368$ &$92.09 \pm 0.340$ & $92.47 \pm 0.360$ & $97.71 \pm 0.160$  \\

BlindNet~\cite{ahn2024style} & $86.95 \pm0.624$ &  $92.63\pm0.356$ & $92.90 \pm 0.330$ & $92.43 \pm 0.350$ & $97.34 \pm 0.190$ \\

RobustSurg (Ours) & $\mathbf{89.32 \pm0.376}$ & $\textbf{93.64}\pm\textbf{0.412}$ &
$\mathbf{96.91 \pm 0.300}$ & $\mathbf{93.89 \pm 0.370}$ & $\mathbf{98.62 \pm 0.140}$   \\

\midrule
Out-of-domain distribution&\multicolumn{5}{c}{HeiCholeSeg Centre 1}\\
\midrule

Baseline \cite{chen2018encoder} & $44.48\pm0.589$ & $58.02\pm0.418$ & $64.47\pm0.395$ & $60.36\pm0.442$ & $76.05\pm0.183$ \\

IBN-Net \cite{pan2018two} & $45.25\pm0.403$ & $58.74\pm0.309$ & $\underline{65.27}\pm\underline{0.284}$ & $62.82\pm0.326$ & $76.56\pm0.159$ \\

RobustNet \cite{choi2021robustnet} & $45.61\pm0.538$ & $58.82\pm0.398$ & $63.65\pm0.363$ & $62.02\pm0.381$ & $75.78\pm0.169$ \\

SNR \cite{jin2021style} & $46.00\pm0.429$ & $59.03\pm0.437$ & $61.74\pm0.389$ & $\underline{68.50}\pm\underline{0.415}$ & $76.87\pm0.172$ \\

SAN-SAW \cite{peng2022semantic} & $39.34\pm0.487$ & $51.96\pm0.466$ & $62.61\pm0.452$ & $56.65\pm0.436$ & $67.52\pm0.208$ \\

SHADE~\cite{zhao2022style} & $\underline{46.37} \pm\underline{0.518}$ &  $\underline{59.40} \pm\underline{0.367}$ & $64.28 \pm 0.330$ & $63.72 \pm 0.340$ & $72.24 \pm 0.110$ \\

SAMed \cite{SAMed} & $36.01\pm0.456$ & $45.89\pm0.345$ & $46.83\pm0.331$ & $51.45\pm0.362$ & $56.12\pm0.165$ \\

BlindNet \cite{ahn2024style} & $46.16\pm0.603$ & $\underline{59.40}\pm\underline{0.339}$ & $64.51\pm0.315$ & $67.78\pm0.344$ & $\underline{76.98}\pm\underline{0.189}$ \\

RobustSurg (Ours) & $\mathbf{48.55\pm0.352}$ & $\mathbf{61.35\pm0.401}$ & $\mathbf{65.49\pm0.289}$ & $\mathbf{73.78\pm0.363}$ & $\mathbf{81.10\pm0.138}$ \\
\midrule
Out-of-domain distribution&\multicolumn{5}{c}{HeiCholeSeg Centre 2}\\
\midrule

Baseline \cite{chen2018encoder} & $40.96\pm0.578$ & $55.50\pm0.389$ & $56.24\pm0.371$ & $74.15\pm0.441$ & $67.20\pm0.183$ \\

IBN-Net \cite{pan2018two} & $44.46\pm0.412$ & $58.52\pm0.344$ & $61.01\pm0.331$ & $\underline{74.40}\pm\underline{0.360}$ & $69.30\pm0.157$ \\

RobustNet \cite{choi2021robustnet} & $\underline{45.63}\pm\underline{0.498}$ & $\underline{59.71}\pm\underline{0.381}$ & $61.30\pm0.352$ & $73.49\pm0.377$ & $67.75\pm0.172$ \\

SNR \cite{jin2021style} & $43.98\pm0.395$ & $58.55\pm0.402$ & $59.81\pm0.375$ & $74.19\pm0.391$ & $75.67\pm0.169$ \\

SAN-SAW \cite{peng2022semantic} & $44.93\pm0.439$ & $58.83\pm0.417$ &  $\underline{65.31}\pm\underline{0.305}$ & $61.33\pm0.357$ & $68.73\pm0.210$ \\
SHADE~\cite{zhao2022style} & $44.73 \pm0.480$ &  $59.32 \pm0.320$ & $56.43 \pm 0.340$ & $73.20 \pm 0.380$ & $65.36 \pm 0.150$ \\
SAMed \cite{SAMed} & $38.11\pm0.461$ & $47.44\pm0.358$ & $51.72\pm0.336$ & $51.79\pm0.369$ & $59.22\pm0.162$ \\

BlindNet \cite{ahn2024style} & $43.92\pm0.567$ & $58.15\pm0.366$ & $61.93\pm0.341$ & $70.57\pm0.359$ & $\underline{75.77}\pm\underline{0.186}$ \\

RobustSurg (Ours) & $\mathbf{50.40\pm0.341}$ & $\mathbf{64.25\pm0.378}$ & $\mathbf{66.95\pm0.297}$ & $\mathbf{73.62\pm0.354}$ & $\mathbf{76.15\pm0.138}$ \\

\bottomrule

\end{tabular}
\end{table*}

\section{Implementation details} \label{sec:implementation_details}

The proposed method was implemented using the PyTorch framework. All the experiments were conducted  using 2 NVIDIA Tesla P100-SXM2-16GB GPUs. The model was trained until convergence was reached as per the early stopping criteria. The SGD optimiser and starting learning rate of $1e^{-2}$ and momentum of 0.9 with polynomial learning rate scheduling with power of 0.9 was used for loss minimization. We also use various augmentations such as color jittering, Gaussian blur, random cropping, random horizontal flipping and random scaling.

\subsection{Training and test setup}
We conducted experiments on three different dataset settings to validate the model performance on both in-domain and out-of-domain distributions. In the first setup, we used the publicly available CholecSeg8K~\citep{hong2020cholecseg8k} for training and in-domain testing and our newly introduced HeiCholeSeg cnter 1 and 2 for evaluation. 
We randomly split CholecSeg8k dataset into 80\%, 10\%, 10\% for training, validation and test, respectively. For unseen centre generalisability test, we used all the annotated images from both centres of HeiCholeSseg centre. In the second setup, we used the EndoUDA dataset for validating  the modality-agnostic performance of our model. We use the standard splits for training, validation and testing reported in~\cite{celik2021endouda} for both the wite light imaging (WLI) and narrow-band imaging (NBI) modalities. In the third setup, we train our model using standard train-val-test splits of publicly available cataract surgery dataset (CaDIS) and for OOD testing, we used cataract-1K dataset which was recorded at a different hospital.

\subsection{Evaluation metrics} Widely used intersection-over-union ($\mathrm{IoU} = \frac{TP}{TP + FP + FN}$), dice similarity coefficient ($\mathrm{DSC} = \frac{2TP}{2TP + FP + FN}$), mean pixel accuracy ($\mathrm{Pixel\ Accuracy} = \frac{TP + TN}{TP + TN + FP + FN}$), precision ($\mathrm{Prec.} = \frac{TP}{TP + FP}$) and recall ($\mathrm{Rec.} = \frac{TP}{TP + FN}$) have been used for comparing the segmentation results of the tested models. Here TP, TN, FP, FN represent true positive, true negative, false positive, false negative respectively.  

\subsection{Comparison with state-of-the-art methods}

DeepLabv3+~\citep{chen2018encoder} with ResNet50 backbone is established as the baseline method in our work. We use subsequent SOTA methods addressing the problem of model generalisability. For instance, we compare with IBN-Net which introduced the idea of instance normalization followed by RobustNet which combined the instance norm with instance whitening. We also carried out a comparison with the style restitution method called SNR, and  other instance norm based methods such as  SAN-SAW and BlindNet which utilise class-wise distributed alignment in the segmentation decoder. Style diversification being a strong approach in DG, we also include comparison with SHADE which proposed style-diversified samples. Finally, we include an SAM-based method to evaluate its performance in the endoscopic domain.

\section{Results and discussion}\label{sec:results-discussion}

In this Section, we comprehensively present results  on different IID and OOD data settings. As our main focus is developing generalisable models on surgical dataset settings, we first discuss quantitative results on both IID and OOD settings in Section~\ref{sec:surgical}. Section~\ref{sec:perclass} provides the results on individual classes for both IID and OOD datasets. Qualitative results on presented in Section~\ref{qualitative}. Afterwards, we discuss ablation experiments by taking CholecSeg8K dataset validation set in Section~\ref{ablations}. In the preceding Section~\ref{sec:other_datasets}, we provide additional experimental results on two other dataset settings, one with cross-centre cataract surgery dataset (CaDIS and Cataract-1K), and one with cross-domain endoscopic dataset (EndoUDA). 
%
% \iffalse
% \hline
% Baseline \cite{chen2018encoder} & $87.67\pm0.12$ & 94.57 & 91.66& 97.78& & $70.52\pm0.14$ & 83.34 & 86.61 & 78.15 \\
% \hline
% IBN-Net  \cite{pan2018two}& $86.82\pm0.13$ & 93.26 & 92.00& 97.76 && $72.43\pm0.16$ & 83.32 & 83.35 & 82.56 \\
% \hline
% RobustNet  \cite{choi2021robustnet}   & $87.67\pm0.12$ & 93.25 & 92.31 & 97.85 && $68.01\pm0.14$ & 76.66 & 83.32 & 85.47\\
% \hline
% SAMed \cite{SAMed} & $86.00\pm0.13$ & 92.09 & 92.47 & 97.71 && $71.40\pm0.18$ & 82.76& 83.35 & 85.99 \\

% \hline
% SNR \cite{jin2021style} & $88.30\pm0.14$ &93.14& 93.84 & 97.90 && $71.28\pm0.16$ &82.52 & 83.33 & 86.71 \\
% \hline

% Rein \cite{wei2024stronger} & 88.45 & & & 93.10 && 38.63 & & & 45.29 \\
% \hline
% BlindNet \cite{ahn2024style} & 86.95 & 92.90 & 92.43 & 92.90 \\ 
% \hline
% RobustSurg (Ours) & $\mathbf{89.32\pm0.12}$ & \textbf{96.91}& \textbf{93.89} &\textbf{98.62} && $\mathbf{75.17\pm0.19}$ &\textbf{86.18} & \textbf{86.65} & \textbf{89.69} \\
% \fi
% In this section, we present quantitative amd qualitative results with DeepLabv3+ \cite{chen2018encoder} as our baseline, IBN-Net \cite{pan2018two}, RobustNet \cite{choi2021robustnet} and RobustSurg which is our proposed method. 

\subsection{Quantitative results} \label{sec:surgical}

% \noindent \textbf{Quantitative results on CholecSeg8K and HeicholeSeg datasets.}
\noindent \textbf{In-domain distribution:} Table~\ref{tab1:results} shows that our method (RobustSurg) achieves SOTA performance on the in-domain CholecSeg test set on all evaluation metric. Specifically, it achieves the highest mean IoU of 89.32  surpassing the the baseline by almost 2\% and the second best method by 1.2\% on mean IoU score. RobustSurg also outperforms other methods on Dice similarity coefficient, for example achieves 1.2\% higher mean DSC as compared to the baseline. Apart from the overlap, it is also important to reduce false positive rate.  Therefore, in terms of precision and recall, RobustSurg achieves 2.5\% higher precision than the second best method which is baseline, and outperforms all other methods on recall score by 1-2\% except SNR which performs almost equally well. Similar trends can be observed on pixel-level performance where RobustSurg gets approximately 1\% higher pixel accuracy as compared to other methods. 

%These results validate the effectiveness of RobustSurg in handling intra-domain variability and delivering reliable semantic segmentation performance under standard evaluation protocols.

\noindent \textbf{Out-of-the domain distribution:}

\noindent To evaluate the robustness and generalisability of segmentation models, we tested on the HeiCholeSeg dataset from two independent clinical centres. As shown in Table~\ref{tab1:results}, our proposed method, RobustSurg, significantly outperforms all competing methods across both domains. On HeiCholeSeg centre 1, RobustSurg achieves the highest mean IoU 48.55 which is 9.17\% higher than the baseline and 4.7\% than the SHADE which is the second best method. Similarly, our method produces a Dice score of 61.35 which is 5.73\% and 3.28\% higher than the baseline and SHADE. Similar trends can be observed on other evaluation metrics. Our method achieves even higher metrics on centre 2 data where it outperforms baseline by 23.04\% IoU score and the second best method RobustNet by 10.45\%. It also achieves 15.76\% and 7.60\% higher Dice score against baseline and second best method. RobustSUrg achieves consistently higher performance on other evaluation metrics compared to all the SOTA methods.

\begin{figure*}[ht!]
    \centering
\includegraphics[width=\linewidth]{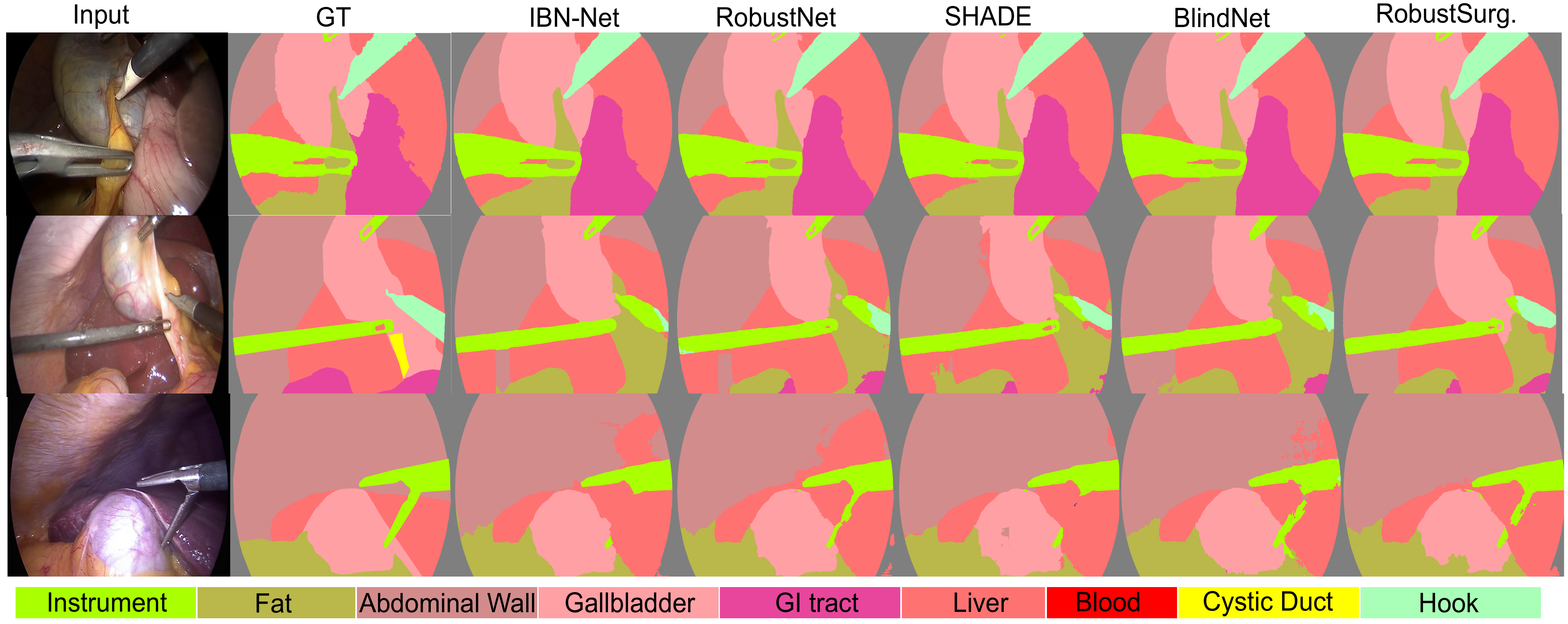}
    \caption{\textbf{Qualitative results.} Top row rows contain qualitative performance on IID CholecSeg8k and bottom two rows show OOD HeiCholeSeg datasets on centre 1 and centre 2, respectively.}
    \label{fig:qualitative}
\end{figure*}

% If we compare various loss functions, ISW gives superior performance. 

% Fig. \ref{fig. qualitative} shows the qualitative results from different models. It is clearly evident that the baseline is not able to segment most of the classes. IBN-Net performs better than the baseline, however, it struggles to segment few classes or contains irregularities in the segmented region. The proposed model with all different loss functions performs almost equally well. It is able to segment most of the regions accurately. However, it fails in some cases: for instance, the Blood (Red) class is not properly segmented in the first row, column 5, 6 and 7. 
 
 % providing slightly higher mIoU of 89.9\% over 87.7\%, 86.8\% and 87.4\% on RobustNet, IBN-Net and DeepLabv3+ respectively.
\subsection{Class-wise performance} \label{sec:perclass}
% \noindent\textbf{Per-class IoU scores}:
\noindent We also report mean pixel accuracies and IoU scores (Table \ref{tab:class}) on every semantic  class on both CholecSeg8k IID and OOD HeiCholeSeg centres 1 and 2. Our method achieved higher pixel accuracy showing robust performance. On IID, the results indicate that our method performs well on most classes, for instance, our method produces higher IoU score on all the classes except blood, hepatic vein, gallbladder and liver ligament where it shows competitive performance. On HeiCholeSeg center 1, RobustSurg produces higher IoU score on liver (69.32 against the second best 64.45 on RobustNet) , grasper instrument (60.23 against the second best 59.86 on RobustNet), and Abdominal wall (68.42 against second best 64.54 on abdominal wall). RobustSurg also produces the second best performance on hook instrument (27.36 against best 29.03 on SHADE), and perform competitively on Gallbladder class (48.20 against the best 50.59 on IBN-Net). Similarly on center 2, RobustSurg achieves 6.7\% higher IoU on liver, 4\% higher IoU on GI tract, and 23.2\% IoU on abdominal wall classes against the second best methods. Further, RobustSurg performance remains the second best on background class and competitive on fat and gallbladder classes.

\subsection{Qualitative results} \label{qualitative}
% \noindent \textbf{Qualitative results}:
\noindent Fig.~\ref{fig:qualitative} presents the qualitative results on both in-domain distribution (row 1),  OOD centre 1 (row 2) and OOD centre 2 (row 3). In the in-domain dataset,  most of the methods are able to distinguish classes correctly; however, the accuracy of segmentation varies. For instance, the grasper instrument on top is more precisely segmented by our method, where other methods struggle to correctly segment it.  In the unseen dataset HeiChole (row 2), methods struggle to distinguish between grasper and hook instrument classes due to the vendor differences between instrument types between training and testing domains. IBN-Net suffers from oversegmentation in the grasper instrument class. In the centre 2 (row 3), most methods struggle on one leg of grasper, and the part above the instrument is misclassified, but our method segments the semantic classes more accurately.

\subsection{Statistical analysis} \label{sec-stat_analysis}
Fig.~\ref{fig:stat_test} demonstrates the paired $t$-test analyses between RobustSurg and other methods on both in-domain and out-of-domain datasets. It can be observed that RobustSurg shows highly significant difference compared to all other methods on all datasets, except with RobustNet on HeiCholeSeg centre 1,  where $p$-value is borderline. RobustSurg also exhibited smaller standard deviation compared to most of the methods, as reflected by narrower interquartile (IQRs). On the CholecSeg data, RobustSurg shows a very narrow deviation as compared to SAMed and SAN-SAW and a competitive performance with other methods. Similarly, on both HeiCholeSeg centres, our method acheives smaller deviation and higher median values as compared to all other methods. These results highlight RobustSurg’s robustness to institutional variability and imaging conditions differences, with statistical significance reached in the vast majority of pairwise comparisons.

\subsection{Computational cost analysis} \label{sec-cost_analysis}

Table~\ref{tab:cost} shows the comparison of computational cost analysis of our method and the state-of-the-art. We report this cost analysis in terms of number of parameters, GFLOPS, and inference time. We can observe that when compared with similar other models such as DeepLabv3+, IBN-Net, RobustNet, and SNR, our method brings a small increase in model size and computational complexity. For example RobustSurg only brings an increase of 1.5\% in model size (45.785M vs. 45.077M), and 0.02\% increase in the computational complexity (436.83 vs. 436.70) and a 5.12\% slower inference as compared to the DeepLabv3+ baseline.

\begin{table*}[!t]
\centering
\caption{Comparison of computational cost. All the models were tested with an image of size 480×860 on NVIDIA Tesla P100-SXM2-16GB GPU. The inference time was averaged over all the validation set.  } 
\label{tab:cost}
\begin{tabular}{lccc}
\toprule
 
Method & \# of Params & GFLOPs & Inference Time (ms)     \\
\midrule

DeepLabv3+~\cite{chen2018encoder} & 45.077M & 436.70 & 11.7   \\
 
IBN-Net~\cite{pan2018two} & 45.079M & 436.84 & 11.9     \\
 
RobustNet~\cite{choi2021robustnet}   &      45.077M  &      436.81  &    11.9   
 \\
 
SNR~\cite{jin2021style} &  46.068M &   459.12  & 13.1  \\
 
SAN-SAW~\cite{peng2022semantic} &  25.069M & 655.25 & 16.1    \\
 
SHADE~\cite{zhao2022style} &  45.077M &  436.70 & 34.7    \\

SAMed~\cite{SAMed} &  639.18M &    5490 & 122.5  \\

BlindNet~\cite{ahn2024style} &  50.100M & 218.40  & 11.2   \\
 
RobustSurg &    45.785M       &   436.83   &   12.3    \\
\bottomrule
\end{tabular}
\end{table*}

% TODO: put colors
% 
%

\subsection{Ablation study} \label{ablations}
In this section, we present different ablation experiments to demonstrate the effectiveness of adding individual components into our proposed RobustSurg method. 

\noindent \textbf{Impact of different backbones.} To validate the wide applicability of RobustSurg, we conducted experiments with two other backbone networks, MobileNetv2 and ShuffleNetv2. Table~\ref{tab-backbones}
shows the results on both in-domain and out-of-domain datasets. Our method shows better performance with both backbones as compared to RobustNet, IBN-Net and baseline with both backbones. This shows that our domain-invariant learning pipeline is backbone-agnostic and can be used with variety of backbone architectures.

\noindent \textbf{Design choices of whitening and restitution blocks.}
Table \ref{tab-design choices} ablates different network configurations where we restrict ISW to first three layers since deeper layers are not rich in style information and RobustNet has shown the best performance on these layers. Afterwards, we change SNR between the residual layers and see how the network behaves. It can be observed that the proposed SRW block (ISW and SNR) configured between layer 1 to layer 3 of ResNet50 produces the best performance.

%\addtocounter{table}{-2}
\begin{table*}[h]
\centering
\caption{Quantitative results for pixel accuracy and IoU for each semantic class. The models are trained on CholecSeg and tested on in-domain CholecSeg test set and HeiCholeSeg Centre 1 and Centre 2 using a ResNet50 backbone. The best and second best results are \textbf{bolded} and \underline{underlined}, respectively.}
\label{tab:class}
\resizebox{\textwidth}{!}{%
\begin{tabular}{l|cc|ccccccccccccc}
\hline
&&&
\cellcolor[RGB]{127,127,127} & 
\cellcolor[RGB]{255,114,114} & 
\cellcolor[RGB]{231,70,156} & 
\cellcolor[RGB]{186,183,75} & 
\cellcolor[RGB]{170,255,0} & 
\cellcolor[RGB]{255,85,0} & 
\cellcolor[RGB]{210,140,140} & 
\cellcolor[RGB]{255,0,0} & 
\cellcolor[RGB]{255,255,0} & 
\cellcolor[RGB]{169,255,184} & 
\cellcolor[RGB]{0,50,128} & 
\cellcolor[RGB]{255,160,165} & 
\cellcolor[RGB]{111,74,0} \\
Methods &
\begin{tabular}[c]{@{}c@{}}Pixel\\ Accuracy\end{tabular} &
mIoU &
\rotatebox[origin=l]{90}{Background} &
\rotatebox[origin=l]{90}{Liver} &
\rotatebox[origin=l]{90}{GI tract} &
\rotatebox[origin=l]{90}{Fat} &
\rotatebox[origin=l]{90}{Grasper} &
\rotatebox[origin=l]{90}{Connective tissue} &
\rotatebox[origin=l]{90}{Abdominal Wall} &
\rotatebox[origin=l]{90}{Blood} &
\rotatebox[origin=l]{90}{Cystic Duct} &
\rotatebox[origin=l]{90}{Hook} &
\rotatebox[origin=l]{90}{Hepatic Vein} &
\rotatebox[origin=l]{90}{Gallbladder} &
\rotatebox[origin=l]{90}{Liver Ligament} \\
\hline \hline
Baseline~\cite{chen2018encoder} & 97.78 & 87.67 & 98.52 & 94.92 
&88.82 & \underline{94.80} &89.46 &88.70 &97.44
&75.21 &75.42 &91.80 &55.91
&93.12 &95.59 \\

IBN-Net~\cite{pan2018two} & 97.76 & 86.82 & 98.82 & 94.92 &
87.18 &
94.45 &
88.43 &
88.77 &
97.45 &
74.31 &
71.33 &
91.67 &
51.76 &
92.76 &
96.82 \\
RobustNet~\cite{choi2021robustnet} & 97.85 & 87.67 & 98.70 &
95.12 &
\underline{90.21} &
94.72 &
89.38 &
89.34 &
97.40 &
73.87 &
72.36 &
92.36 &
55.32 &
\textbf{93.76 }&
\textbf{97.84} \\
SNR~\cite{jin2021style} & \underline{97.90} & \underline{88.30} & \underline{99.01} & \underline{95.15} & 88.82 & 94.65 & \underline{89.54} & 88.75 & \underline{97.60} & \textbf{76.88} & \underline{78.09} & \underline{92.45} & 56.57 & 93.28 & 97.42 \\

SAN-SAW~\cite{peng2022semantic} & 96.16 & 82.54 & 98.95 & 
91.96 & 
83.69 & 
92.04 & 
82.75 & 
78.09 & 
93.12 & 
63.80 & 
70.77 & 
86.58 & 
53.51 & 
85.79 & 
91.97 \\
SHADE~\cite{zhao2022style}  & 97.60 & 86.78 & 97.90 &
94.50 &
87.40 &
94.60 & 
88.00 &
88.60 &
97.30 &
71.40 &
72.40 &
90.30 &
58.50 &
93.00 &
94.30\\
SAMed~\cite{SAMed} & 97.71 & 86.00 & 96.14 & 
93.57 & 
84.48 & 
92.37 & 
80.36 & 
\underline{88.94} & 
93.80 & 
67.01 & 
76.43 & 
89.25 & 
\textbf{68.54} & 
92.40 & 
94.73\\

BlindNet~\cite{ahn2024style} & 97.34 & 86.42 & 98.04 &
94.33 &
87.44 &
94.57 &
87.29 &
88.14 &
96.98 &
71.16 &
72.12 &
90.15 &
55.60 &
92.48 &
95.15 \\
RobustSurg. (ours) & \textbf{98.62} & \textbf{89.32} & \textbf{99.42} & 
\textbf{97.83} & 
\textbf{92.52} & 
\textbf{95.32} &
\textbf{91.97} & 
\textbf{91.32} & 
\textbf{98.36} & 
\underline{75.95} & 
\textbf{79.82} & 
\textbf{92.46} & 
\underline{62.31} & 
\underline{92.92} &
\underline{97.79} \\

\hline
\textbf{HeiCholeSeg Centre 1} \\
\hline

Baseline~\cite{chen2018encoder} & 76.05 & 44.48 & 82.10 & 
63.27 & 
17.73 & 
\underline{15.59} &  
51.93 & -- &  
58.92 &  -- & -- &
20.39 & -- &
45.98 & --   \\

IBN-Net~\cite{pan2018two} & 76.56 & 45.25 & 85.70 & 
63.52 & 
\underline{21.52} & 
14.79 & 
54.26 & -- & 
52.06 & -- & -- &
19.58 & --& 
\textbf{50.59 }& -- \\
RobustNet~\cite{choi2021robustnet} & 75.78 & 45.61 & 83.88 & 
\underline{64.45} & 
16.42 & 
13.16 & 
\underline{59.86}  & -- & 
60.01 &  -- & -- & 
26.22 & --& 
40.94 & -- \\
SNR~\cite{jin2021style} & 76.87 & 46.00 & \textbf{93.91} & 
58.22 & 
\textbf{53.63} & 
\textbf{35.25} & 
50.36 & -- & 
34.50 &  -- & -- & 
23.40 & -- & 
18.79 & --  \\

SAN-SAW~\cite{peng2022semantic} & 67.52 & 39.34 & \underline{87.67} & 
54.22 &
11.57 &
9.65 &
55.02 & -- & 
44.59 & -- & -- & 
17.00 & -- & 
35.00 & --\\
SHADE~\cite{zhao2022style}  & 72.24 & 46.37 & 85.40 &
61.23 &
15.01 &
13.23 &
58.02 & -- &
\underline{64.54} & -- & -- & 
\textbf{29.03} & -- &
44.52 & --\\
SAMed~\cite{SAMed} & 56.12 & 36.01 & 64.24 & 
61.16 & 
14.00 & 
9.13 & 
34.26 & -- & 
45.25 &  -- & -- & 
10.40 &  --& 
\underline{49.62} & --\\

BlindNet~\cite{ahn2024style} & \underline{76.98} & \underline{46.16} & 84.39 & 
63.03 & 
19.00 & 
10.95 & 
59.29 & --& 
58.51 & -- & -- &
26.42 &  --& 
47.76 & -- \\
RobustSurg. (ours) & \textbf{81.10} & \textbf{48.55} & 84.79 & 
\textbf{69.32} & 
20.57 & 
10.95 & 
\textbf{60.23 }& -- & 
\textbf{68.42} &  -- & -- &
\underline{27.36} & -- &
47.38 & -- \\

\hline
\textbf{HeiCholeSeg Centre 2} \\
\hline

Baseline~\cite{chen2018encoder} & 67.20 & 44.48 & 87.89 & 
52.54 & 
34.35 & 
\textbf{37.93 }& 
44.09 & -- & 
27.13 & -- & -- &
17.90 & -- &
25.92 & --   \\

IBN-Net~\cite{pan2018two} & 69.30 & 44.46 & 93.47 & 
61.58 & 
49.18 & 
\underline{36.28} & 
42.68 & -- & 
29.30 &  -- & -- &
23.75 &  -- & 
19.49 & --\\
RobustNet~\cite{choi2021robustnet} & 67.75 & \underline{45.63} & 93.69 & 
56.12 & 
\underline{52.60} & 
34.70 & 
50.95 & -- &  
29.00 & -- & -- &
27.60 & -- &
20.39 &  \\
SNR~\cite{jin2021style} & 75.67 & 43.98 & 91.16 & 
56.71 & 
39.88 & 
35.80 & 
49.17 & --& 
27.21 & -- & -- &
29.23 & -- &
22.73 & -- \\

SAN-SAW~\cite{peng2022semantic} & 68.73 & 44.93 & \textbf{96.88 }&
61.09 & 
22.71 & 
32.56 & 
\underline{52.89} & -- & 
39.82 & -- & -- & 
\textbf{30.53} & -- &
22.94 & --\\
SHADE~\cite{zhao2022style}  & 65.36 & 44.73 &  
91.50 & 
51.60 & 
32.50 & 
34.20 & 
\textbf{56.70} & -- & 
36.40& -- & -- &
20.10& -- &
\underline{34.90} & --\\
SAMed~\cite{SAMed} & 59.22 & 38.11 & 75.96 & 
62.40 & 
14.43 & 
25.23 & 
33.25 & -- & 
\underline{41.53} & -- & -- &
16.94 & -- &
\textbf{35.14} & --\\

BlindNet~\cite{ahn2024style} & \underline{75.77} & 43.92 & 92.96 & 
\underline{63.86} & 
31.58 & 
31.33 & 
46.27 & -- &
34.93 & -- & -- &
\underline{29.70} & -- &
20.78 & --\\
RobustSurg. (ours) & \textbf{76.15} & \textbf{50.40} & \underline{96.33} &
\textbf{68.13} & 
\textbf{54.63} & 
31.99 & 
49.45 & -- & 
\textbf{51.17} & -- & -- &
23.63 & -- & 
27.92 & --\\

\hline
\end{tabular}%
}

\end{table*}

\begin{figure*}[hbt!]
    \centering
\includegraphics[width=1\linewidth]{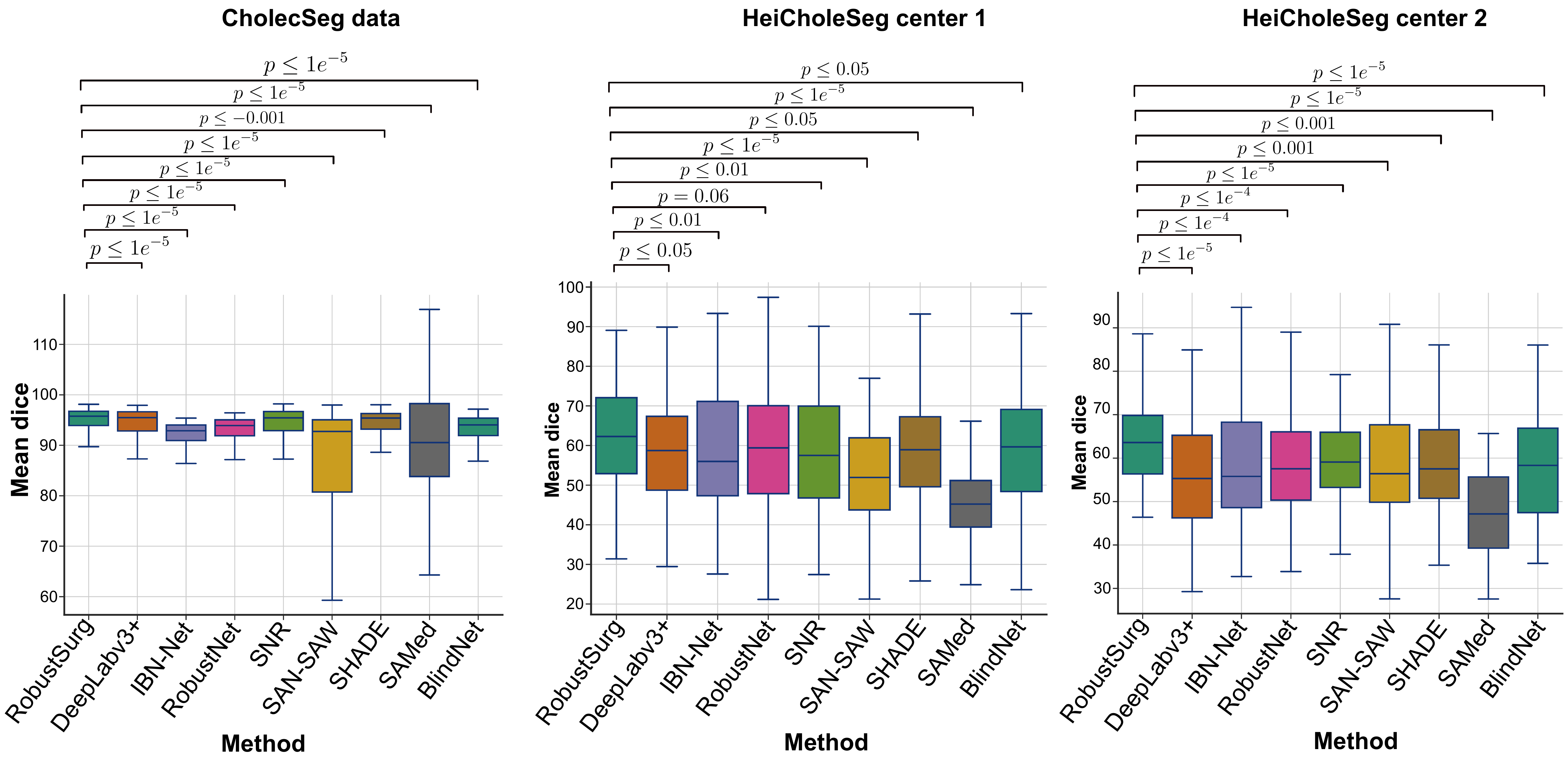}
    \caption{Box-plots for paired t-test on in-domain CholecSeg dataset and OOD HeiCholecSeg centre 1 and centre 2 datasets.}
    \label{fig:stat_test}
\end{figure*}

\begin{table*}[ht]
\centering
\caption{Comparison of IoU and Dice scores for different methods under MobileNet and ShuffleNet backbones across three datasets. The best and second best results are \textbf{bolded} and \underline{underlined}, respectively.}
\label{tab-backbones}
\begin{tabular}{llcccccc}
\toprule
Backbone & Method & \multicolumn{2}{c}{CholecSeg8K} & \multicolumn{2}{c}{HeiCholeSeg-C1} & \multicolumn{2}{c}{HeiCholeSeg-C2} \\
 & & IoU & Dice & IoU & Dice & IoU & Dice \\
\midrule
\multirow{3}{*}{MobileNetv2} 
  & Baseline \cite{chen2018encoder} & 87.12 & 92.64 & 44.71 & \underline{58.13} & 34.46 & 50.37 \\
  & IBN-Net  \cite{pan2018two} & \underline{87.32} & \underline{92.81} & 44.63 & 57.56 & \underline{45.12} & \underline{59.15} \\
  & RobustNet  \cite{choi2021robustnet} & 85.73 & 91.71 & \underline{44.88} & 58.10 & 42.78 & 57.55 \\
   & RobustSurg (Ours)   & \textbf{87.95} & \textbf{92.88} & \textbf{47.91} & \textbf{61.15} & \textbf{50.11} & \textbf{63.99} \\
\midrule
\multirow{3}{*}{ShuffleNetv2} 
  & Baseline~\cite{chen2018encoder} & 86.41 & 92.19 & 44.25 & 57.25 & 44.71 & 58.76 \\
  & IBN-Net~\cite{pan2018two} & 86.80 & 92.50 & \underline{45.85} & \underline{58.50} & 45.05 & 59.18 \\
  & RobustNet~\cite{choi2021robustnet} & \underline{86.98} & \underline{92.77} & 44.20 & 57.27 & \underline{46.87} & \underline{61.01} \\
  & RobustSurg (Ours)   & \textbf{87.52} & \textbf{93.20} & \textbf{48.21} &\textbf{ 61.20} & \textbf{50.25} & \textbf{64.01} \\
\bottomrule
\end{tabular}
\end{table*}

\begin{comment}
\begin{table*}[hbt!]
\centering
\caption{Effect of using SNR and ISW blocks in different arrangements between the layers. Layers 1 to 5 represent ResNet-50 layers.  }\label{ab-1}

%\begin{tabular}{cccccccccc}
\begin{tabular}{p{0.8cm}p{0.8cm}p{0.8cm}p{0.8cm}p{0.8cm}p{0.8cm}p{0.8cm}p{0.8cm} p{0.8cm}p{0.8cm}p{0.8cm}p{0.8cm}}

\multicolumn{5}{c}{\textbf{ISW}} & \multicolumn{5}{c}{\textbf{SNR}}\\
 \cmidrule(lr){1-5}  % trim from left/right
\cmidrule(lr){6-10} 
$L_1$ & $L_2$ & $L_3$ & $L_4$ & $L_5$& $L_1$ & $L_2$ & $L_3$ & $L_4$ & $L_5$& \textbf{Test (IID)} & \textbf{Target (OOD)}\\
\hline
\cmark & \cmark& \cmark & \xmark & \xmark& \cmark & \cmark & \cmark& \xmark& \xmark& 90.9 & 75.3\\

\hline
\cmark & \cmark& \cmark & \xmark & \xmark & \xmark& \cmark & \cmark& \cmark & \xmark& 82.9 & 79.3\\

\hline
\xmark & \cmark& \cmark & \cmark & \xmark& \xmark & \cmark & \cmark& \cmark & \xmark& 84.3 & 79.7\\
\bottomrule

\end{tabular}
\end{table*}
\end{comment}

\begin{table*}[]
\centering
\caption{Effect of using single or multiple SNR blocks with fixed ISW blocks. Layers 1 to 5 represent ResNet-50 layers. Best values are represented in \textbf{Bold} and second best \underline{underlined}.   }\label{ab-2}
\label{tab-design choices}
%\begin{tabular}{cccccccccc}
\begin{tabular}{p{0.8cm}p{0.8cm}p{0.8cm}p{0.8cm}p{0.8cm}p{0.8cm}p{0.8cm}p{0.8cm} p{0.8cm}p{0.8cm}p{1.5cm}}
\toprule
\multicolumn{5}{c}{\textbf{ISW}} & \multicolumn{5}{c}{\textbf{SNR}}\\
 \cmidrule(lr){1-5}  % trim from left/right
\cmidrule(lr){6-10} 
$L_1$ & $L_2$ & $L_3$ & $L_4$ & $L_5$ & $L_1$ & $L_2$ & $L_3$ & $L_4$ & $L_5$ & \textbf{Validation (IID)})\\
\midrule
\cmark & \cmark& \cmark & \xmark & \xmark & \xmark & \cmark& \xmark& \xmark & \xmark & 87.82  \\

 \cmark & \cmark& \cmark & \xmark & \xmark & \xmark & \cmark& \cmark& \xmark & \xmark & \underline{88.12}  \\

\cmark & \cmark& \cmark & \xmark & \xmark & \xmark & \cmark& \cmark& \cmark & \xmark & \textbf{88.92}  \\

\cmark & \cmark& \cmark & \xmark & \xmark & \xmark & \xmark& \cmark& \cmark & \xmark & 87.54  \\

\cmark & \cmark& \cmark & \xmark & \xmark & \xmark & \xmark& \cmark& \xmark & \cmark & 87.68  \\
\bottomrule
\end{tabular}
 
\end{table*}

\noindent \textbf{Separating covariance elements.}
We adopt k-means clustering approach to separate the covariance elements into two groups, i.e., domain-dependent styles and domain-invariant content based on the variance of covariance elements of original and photometric transformed images. As discussed in Section~\ref{isw}, we divide the variance elements into k clusters based on variance magnitude. The first m elements represent domain-invariant content (low variance) and the remaining elements are considered sensitive to photometric transformation meaning they represent the domain-specific styles. Table \ref{tab:k value ablation} shows that the optimal k value is 2.

% \begin{table*}[!t]
% % \centering
% \caption{Effect of k value on model performance. Best values are represented in \textbf{Bold} and second best \underline{underlined}.   }\label{ab-3}
% \label{tab:k value ablation}
% \begin{tabular}{lllllllll}
% \toprule
% \multicolumn{1}{c}{} & \multicolumn{4}{c}{\textbf{Test(IID)}} & \multicolumn{4}{c}{\textbf{Test(OOD)}}\\
% \cline{2-5} \cline{6-9} 

% \textbf{Method} & \textbf{IoU} & \textbf{Prec.} & \textbf{Rec.} & \textbf{Acc.}  & \textbf{IoU} & \textbf{Prec.} & \textbf{Rec.} & \textbf{Acc.} \\
% \hline 

% DeepLabv3+ & 90.40 & 96.67 & 94.90 & 91.73 & 60.70	& 70.52	& 72.74 & 	65.45 \\
% \hline
% RobustSurg (\textbf{k=2}) &  88.92 &	94.61	& 93.66	 & 91.96    & 78.80 &	95.91	& 81.54 &	84.53 \\
% \hline
% RobustSurg (k=3) & 88.49 &	96.31 &	94.44 &	91.66
% & 75.32	& 85.02	& 86.84	& 83.30\\
% \hline
% RobustSurg (k=5) & 88.49	& 93.37 & 92.13 &		91.66 &  77.41 &	86.48	& 88.06	& 84.87  \\
% \hline
% RobustSurg (k=7) &  88.38 &	93.50 &	94.17	& 91.62 & 78.18 &	89.39 &	87.39 &	85.91 \\
% \hline
% RobustSurg (k=10) & 87.67 &	94.16	& 92.71	& 90.98 & 78.93	& 89.97	& 86.54	& 85.60 \\
% \hline
% RobustSurg (k=20) & 88.62 &	93.58	& 94.35	& 91.81 & 77.25 &	85.09 &	89.34 &	84.98 \\
% \bottomrule
% \end{tabular}
% \end{table*}

\begin{table}[!t]
% \centering
\caption{Effect of k value on model performance. Best values are represented in \textbf{Bold} and second best \underline{underlined}.   }\label{ab-3}
\label{tab:k value ablation}
\begin{tabular}{lllll}
\toprule
\multicolumn{1}{c}{} & \multicolumn{4}{c}{Validation (IID)}\\
\cline{2-5}  

Method & mIoU & Prec. & Rec. & Acc.    \\
\midrule

DeepLabv3+ & 87.40 & 92.67 & 92.90 & 90.73   \\
 
RobustSurg (\textbf{k=2}) &  \textbf{88.92} &	\textbf{94.61}	& \textbf{94.66}	 & \textbf{91.96}     \\
 
RobustSurg (k=3) & 88.49 &	\underline{94.31} &	\underline{94.44} &	91.66
 \\
 
RobustSurg (k=5) & 88.49	& 93.37 & 92.13 &		91.66      \\
 
RobustSurg (k=7) &  88.38 &	93.50 &	94.17	& 91.62    \\
 
RobustSurg (k=10) & \underline{87.67} &	94.16	& 92.71	& 90.98    \\
 
RobustSurg (k=20) &  88.62 &	93.58	& 94.35	& \underline{91.81}  \\
\bottomrule
\end{tabular}
\end{table}

\noindent \textbf{Effectiveness of dual causality loss.}
Table \ref{tab:dual causality loss} shows the ablation experiments on the dual causality loss constraint. We observe that inclusion of the dual causality loss produces the best performance on both IID and OOD data. The addition of $\mathcal{L}_{dc}$ surpassed the architecture without it by approximately 2\% on IID data and 1.5\% on OOD data. The dual causality loss enhances the process of disentanglement of relevant and non-relevant features. Furthermore, the addition of individual loss components ($ \mathcal{L}_{SNR}^+ and \mathcal{L}_{SNR}^-$) shows that skipping the contaminated features from the overall loss constraint degrades the performance. 

% \begin{table*}[t:]
% \centering
% \caption{Effect of dual causality loss ($\mathcal{L}_{dc}$ = $\mathcal{L}_{SNR}^+$ + $\mathcal{L}_{SNR}^-$).  Best values are represented in \textbf{Bold} and second best \underline{underlined}. }\label{ab-4}
% \label{tab:dual causality loss}
% \begin{tabular}{lllllllll}
% \toprule
% \multicolumn{1}{c}{} & \multicolumn{4}{c}{\textbf{Validation(IID)}} & \multicolumn{4}{c}{\textbf{Validation(OOD)}}\\
% \cline{2-5} \cline{6-9} 
% \textbf{Method} & \textbf{mIoU} & \textbf{Prec.} & \textbf{Rec.} & \textbf{mAcc.} & \textbf{mIoU} & \textbf{Prec.} & \textbf{Rec.} & \textbf{mAcc.}\\
% \hline 
% DeepLabv3+ & 90.40 & 96.67 & 94.90 & 91.73 & 60.70	& 70.52	& 72.74 & 	65.45 \\
% \midrule
% RobustSurg (w/o $\mathcal{L}_{SNR}^+$) & 88.10 & 93.38	& 93.98 &	91.41  & 78.15  &	87.18   &	84.27 	& 83.80   \\
% \midrule
% RobustSurg (w/o $\mathcal{L}_{SNR}^-$) & 87.77 & 94.21 & 91.52 & 90.32 & 76.32  &	87.24   &	82.77  	& 82.02   \\

% \midrule
% RobustSurg (w/o $\mathcal{L}_{dc}$) & 86.41 & 92.58 & 89.45 & 88.98 & 77.25   &	86.83   &	81.37  	& 83.46  \\
% \midrule
% RobustSurg (with $\mathcal{L}_{dc}$) & 88.92 &	94.61	& 93.66	 & 91.96    & 78.80 &	95.91 &	81.54	&  84.53 \\

% \bottomrule
% \end{tabular}
% \end{table*}

\begin{table}[t:]
\centering
\caption{Effect of dual causality loss ($\mathcal{L}_{dc}$ = $\mathcal{L}_{SNR}^+$ + $\mathcal{L}_{SNR}^-$).  Best values are represented in \textbf{Bold} and second best \underline{underlined}. }\label{ab-4}
\label{tab:dual causality loss}
\begin{tabular}{lllll}
\toprule
\multicolumn{1}{c}{} & \multicolumn{4}{c}{Validation (IID)}\\
\cline{2-5} 
Method & mIoU & Prec. & Rec. & Acc. \\
\midrule
DeepLabv3+ & 87.40 & 92.67 & 92.90 & 90.73  \\
 
RobustSurg (w/o $\mathcal{L}_{SNR}^+$) & \underline{88.10} & 93.38	& \underline{93.98} &	\underline{91.41}    \\
 
RobustSurg (w/o $\mathcal{L}_{SNR}^-$) & 87.77 & \underline{94.21} & 91.52 & 90.32   \\

RobustSurg (w/o $\mathcal{L}_{dc}$) & 86.41 & 92.58 & 89.45 & 88.98   \\
 
RobustSurg (with $\mathcal{L}_{dc}$) & \textbf{88.92} &	\textbf{94.61}	& \textbf{94.66}	 & \textbf{91.96}  \\

\bottomrule
\end{tabular}
\end{table}

\begin{figure*}[hbt!]
    \centering
\includegraphics[width=0.6\linewidth]{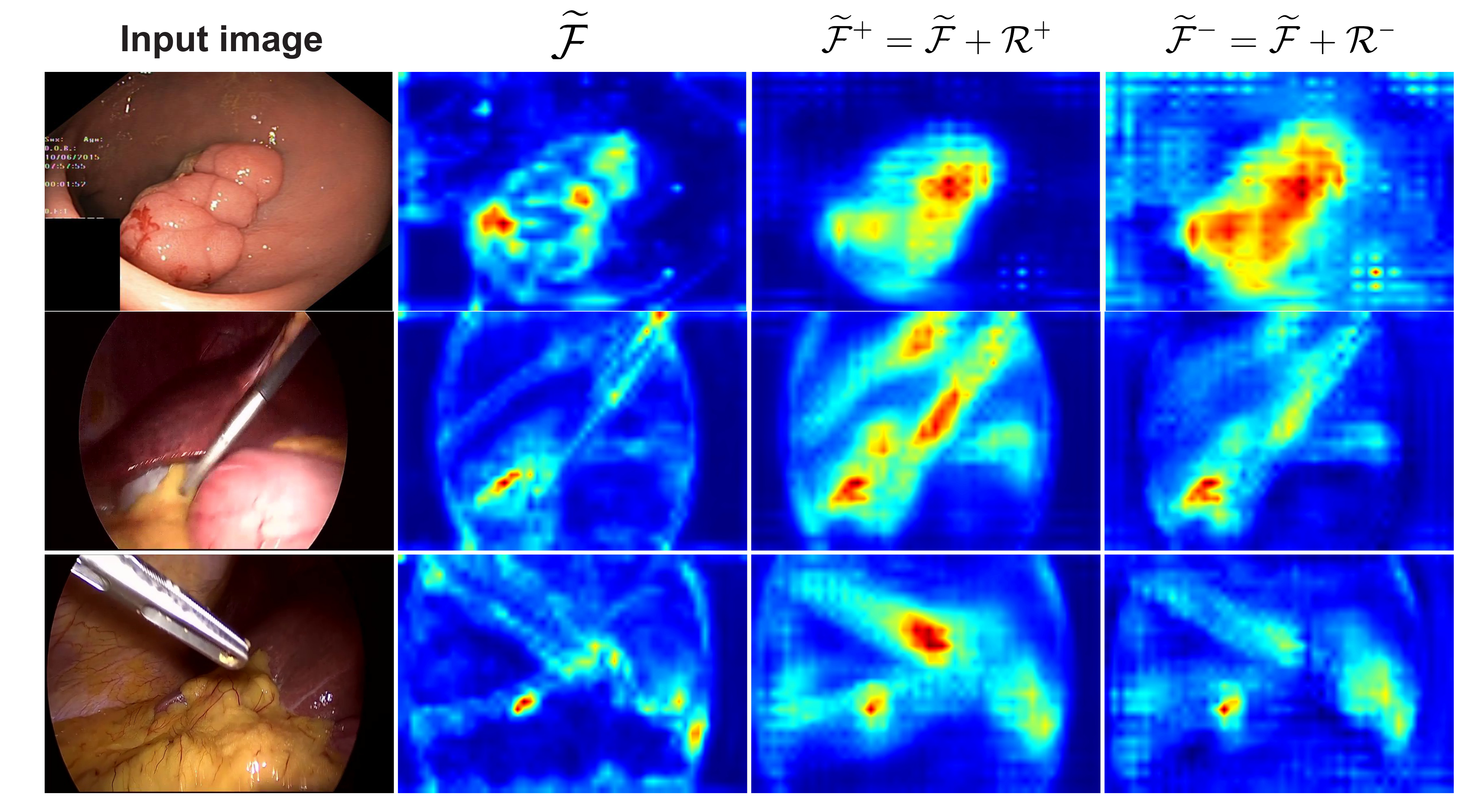}
    \caption{Activation feature maps from the third SNR block: The activation feature maps show that the model can disentangle task-relevant features from the irrelevant features using the residual feature $R^+$. }
    \label{fig:feat_maps}
\end{figure*}

\noindent \textbf{Feature-map visualization.}
 Fig. \ref{fig:feat_maps} presents the feature maps obtained from third SNR block (i.e., SNR-3) on the EndoUDA polyp dataset to understand the effectiveness of SNR block. We obtained individual feature maps by summarizing the map on channel dimension followed by $\mathscr{l}_2$ normalization. Fig. \ref{fig:feat_maps} depicts the original normalised feature map, enhanced features and the contaminated feature map. We can see that addition of the task-relevant feature enhances the model's ability to focus on discriminative areas of the image such as anatomical regions, instruments etc, while addition of task-irrelevant features does not have a uniform response on the discriminative areas.

\begin{figure*}[hbt!]
    \centering
\includegraphics[width=0.6\linewidth]{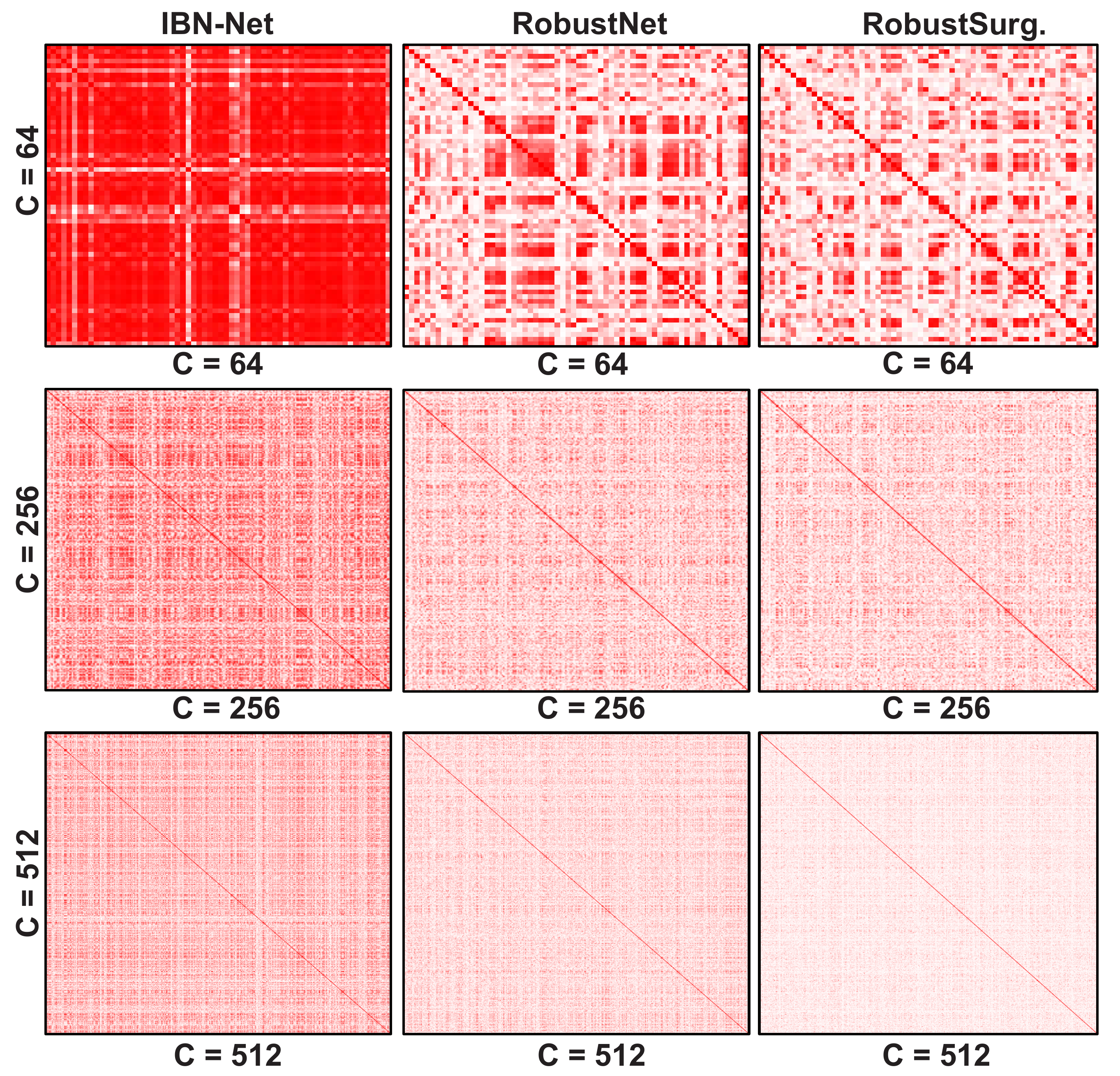}
    \caption{Visualization of covariance matrices extracted from IBN- Net, RobustNet and our model. The first, second and third rows represent feature covariance matrices extracted from the first, the second and third convolution layers, respectively. }
    \label{fig:covariance}
\end{figure*}

\noindent \textbf{Comparison of covariance matrices.}
To demonstrate the enhanced feature whitening operation offered by our approach, we present
different covariance matrices of intermediate feature maps of IBN-Net, RobustNet and our RobustSurg. As shown in Fig. \ref{fig:covariance}, the first row represents covariance matric for first convolutional layer, and second and third rows represent the covariance matrices for second and third convolutional layers respectively. As pointed out by IBN-Net, the style information is encoded in the early layers of the network. We can observe that our method using enhanced features by restitution operation is capable of perming whitening operation to remove domain styles more effectively as compared to other two methods. 

% \section{Results and Discussion} \label{sec:results-discussion}

%TODO: Add a discussion section please!

\subsection{Generalisability tests on endoscopic and cataract surgery datasets.} \label{sec:other_datasets}

\noindent \textbf{Results on EndoUDA (polyp)}: Table~\ref{tab:endoudaquant} shows results on EndoUDA dataset. 
On EndoUDA polyp with IID, our proposed model got 81.67\% mIoU score which is slightly higher than the baseline and SNR method and  5.9\%, 4.4\%, 2.9\% and 4.4\% higher as compared to IBN-Net, RobustNet, SAMed, and BlindNet. On the OOD side, our model with 78.30\% mIoU outperformed the baseline by 21.8\% and other SOTA methods with a significant margin of 11.4\% RobustNet, 15\% IBN-Net, 5.6\% SAMed, 9.7\% SNR, and 46\% BlindNet respectively.  On the mDSC score, similar performance can be observed with our model outperforming the baseline by 22.11\% and second best method SAMed by 8.3\%. 

\noindent \textbf{Results on EndoUDA (BE)}: 
 On EndoUDA (BE) IID, our model produced 91.50 mIoU and 95.01 mDSC score which is higher than the baseline and most of the SOTA methods. RobustSurg outperformed baseline by 3.74\%, and 4.01\% in terms of mIoU and mDSC scores. 
 Similarly, our model surpassed other SOTA methods by a considerable margin. BlindNet performed slightly higher on mIoU and mDSC with 91.93\% and 95.79\% respectively. 
 On the OOD setting, our method performed much better with mIoU score of 86.91\% and 93.00\% which is 43\% and 38\% higher than the baseline on both mIoU and mDSC scores both, and 2\% and 1.1\% better than BlindNet method on the same metrics.  
 
\begin{figure*}[t!]
    \centering
\includegraphics[width=\linewidth]{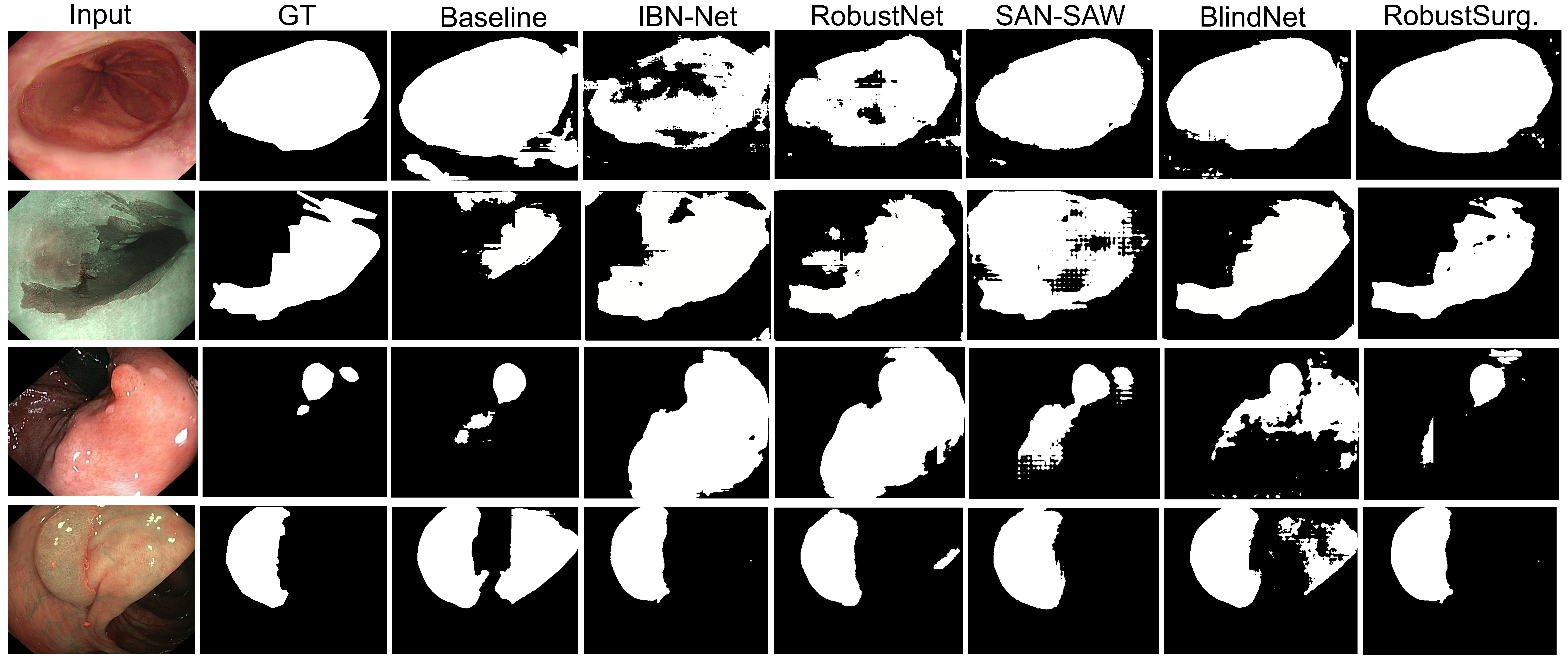}
    \caption{Qualitative results on EndoUDA BE and Polyp datasets. Top two rows contain qualitative performance on WLI and NBI BE images respectively while third and fourth rows contain WLI and NBI modality polyp images.}
    \label{fig:qualendouda}
\end{figure*}

Fig.~\ref{fig:qualendouda} shows the qualitative performance of different methods on EndoUDA Barret's and polyp datasets. It can be observed that RobustSurg is able to achieve clearer segmentation boundaries and can work well on small as well as large polyps.

 \noindent\textbf{Results on CaDIS and Cataract-1K Benchmark}: Table~\ref{tab:cadisquant} presents the quantitative performance of SOTA and our method on Cataract surgery datasets. On the CaDIS test set, our proposed model outperforms the baseline by 1.6\% on mIoU score and 2.8\% on mDSC score, and achieves better results  than most of the SOTA method, for instance 1.23\% higher mIoU, 1.6\% higher mDSC score on second best method. On the Cataract-1K OOD dataset, our proposed model outperforms the baseline by 51.2\% and 35.3\% in terms mIoU and mDSC scores. Our method achieves 3.9\% higher mIoU and 8.2\% higher mDSC as compared to the second best method (BlindNet). 

\begin{figure*}[t!]
    \centering
\includegraphics[width=\linewidth]{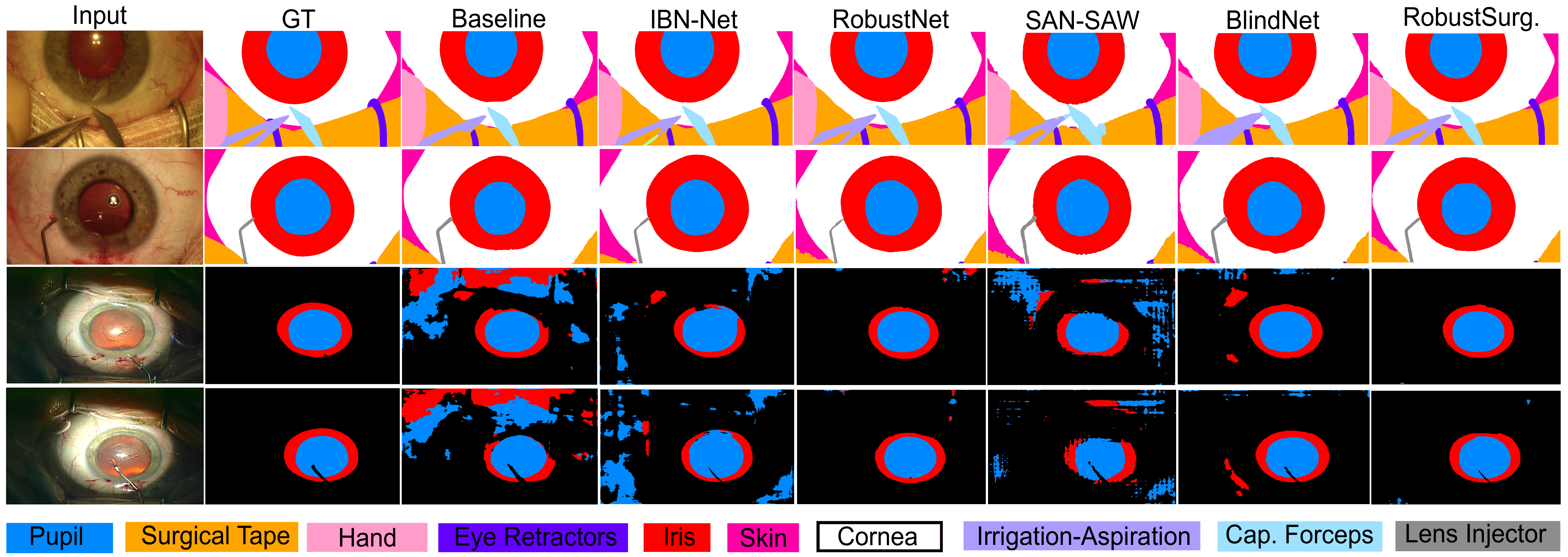}
    \caption{Qualitative results on CaDIS and cataract-1K datasets. Top two rows depict the qualitative performance on CaDIS and Cataract-1K data images respectively.}
    \label{fig:cadisqual}
\end{figure*}

 Fig.~\ref{fig:cadisqual} presents the qualitative performance on in-domain and out-of-domain cataracts datasets. Most method work well in segmenting anatomical landmarks on the CadIS dataset, but struggle to accurately segment instrument classes such as Irrigation-Aspiration. However, our method is able to provide more accurate segmentation boundaries for all the classes. Results on the Cataract-1K dataset are only shown for two classes, where our method can perform better segmentation of Iris and Pupil, but other methods suffer from over-segmentation.

\begin{table*}[t!]
\footnotesize
\caption{Table showing mean Intersection over Union (mIoU), Dice, Precision, Recall and accuracy scores for the SOTA and our proposed method on EndoUDA dataset. Best values are represented in \textbf{Bold} and second best \underline{underlined}. } \label{tab:endoudaquant}
\begin{tabular}{lcccc|cccc}
% \textbf{EndoUDA (polyp)}&\multicolumn{2}{c}{\textbf{Source Modality}}& \multicolumn{1}{c}{\textbf{Target Modality}}\\
\toprule
EndoUDA (polyp)&\multicolumn{4}{c}{Test (IID)}& \multicolumn{4}{c}{Target (OOD)}\\
\midrule
Method & mIoU & mDSC &  Prec. & Rec.  &   mIoU &  mDSC & Prec. & Rec.  \\
\midrule
Baseline \cite{chen2018encoder}   & $\underline{81.42}\pm\underline{0.551}$ & $\underline{90.32}\pm\underline{0.234}$& \underline{90.21} & 90.11 & $64.28\pm0.350$ & $73.45\pm0.234$& 77.88 & \underline{76.84} \\
 
IBN-Net \cite{pan2018two}  & $77.08\pm0.671$ & $87.45\pm0.534$& 89.01 & 85.59 &  $68.04\pm0.230$ & $76.21\pm0.452$& 86.34& 75.63   \\
 
RobustNet \cite{choi2021robustnet}   & $78.18\pm0.310$ & $89.33\pm0.457$& \textbf{90.46} & 85.51 & $70.24\pm0.871$ & $78.34\pm0.332$&\textbf{ 91.18} & 75.65  \\
 
SNR \cite{jin2021style} & $81.06\pm0.176$ & $89.31\pm0.421$& 88.00 & \underline{91.13}  & $71.37\pm0.120$ & $80.40\pm0.321$& 88.72 & 77.89  \\

SAN-SAW \cite{peng2022semantic}  & $74.59\pm0.325$ & $85.45\pm0.347$ & 86.12 & 84.79  & $46.04\pm0.412$ & $63.05\pm0.383$ & 67.57 & 59.10  \\

 SAMed \cite{SAMed}  &  $79.32 \pm0.231$ & $88.45\pm0.356$& 88.42 & 83.76   & $\underline{74.12}\pm\underline{0.154} $ & $\underline{82.78}\pm\underline{0.347}$& 89.30 & 72.55  \\

BlindNet \cite{ahn2024style} & $78.20\pm0.110$ & $87.77\pm0.630$ & 86.72 & 88.84  & $53.49\pm0.237$ & $69.70\pm0.161$ & 78.94 & 62.40  \\
 
RobustSurg (Ours)  & $\textbf{81.67}\pm\textbf{0.871}$ & $\textbf{91.54}\pm\textbf{0.243}$& 89.22 & \textbf{91.32}  & $\mathbf{78.30}\pm\textbf{0.111}$ & $\textbf{89.69}\pm\textbf{0.321}$& \underline{89.96} & \textbf{86.93} \\
\midrule
%%%%%%%%%%%%%%%%%%%%%%%%%%%%%%%%%%%%%%%%%%%%%%%%%%%%%%%%%%%%%%%%%%%%%%%%%%%%%%%%%%%%%%%%%
EndoUDA (BE)& & \\
\midrule
Baseline \cite{chen2018encoder} & $88.20\pm0.764$ & $91.34\pm0.521$&93.87 & 93.63 &   $60.73\pm0.410$ & $67.23\pm0.621$& 70.21 & 65.82  \\
 
IBN-Net \cite{pan2018two}   & $76.82\pm0.842$ & $80.41\pm0.440$& 82.35 & 85.21  & $68.38\pm0.421$ & $75.82\pm0.460$& 80.31 &73.62  \\
 
RobustNet \cite{choi2021robustnet} &  $84.23\pm0.621$ & $88.39\pm0.342$& 91.45 & 90.61   & $71.36\pm0.341$ & $81.90\pm0.42$& 81.57 & 75.54  \\
 
SAMed \cite{SAMed} & $86.88\pm0.463$ & $92.25\pm0.419$ & 93.96 & 91.61 & $75.20\pm0.538$ & $82.75\pm0.503$ & 79.36 & 85.70\\
 
SNR \cite{jin2021style} & $85.37\pm0.333$ & $89.49\pm0.478$ &92.54 & 91.73  & $72.87\pm0.551$ & $83.43\pm0.320$& 84.24 & 76.72   \\

SAN-SAW \cite{peng2022semantic} & $87.59\pm0.387$ & $93.39\pm0.352$ & 93.52 & 93.25 &            $70.53\pm0.346$ & $82.71\pm0.421$ & 78.00 & 84.30   \\

BlindNet \cite{ahn2024style} & $\textbf{91.93}\pm\textbf{0.540}$ & $\textbf{95.79}\pm\textbf{0.301}$  & \underline{96.03} & \textbf{95.56} &  
$\underline{85.15}\pm\underline{0.420}$ & $\underline{91.98}\pm\underline{0.251}$ & \underline{91.47} & \textbf{92.49}   
\\
 
RobustSurg (Ours) & $\underline{91.50}\pm\underline{0.530}$ & $\underline{95.01}\pm\underline{0.409}$ & \textbf{97.06} & \underline{93.35} &   $\textbf{86.91}\pm\textbf{0.461}$ & $\textbf{93.00}\pm\textbf{0.273}$ & \textbf{95.08} & \underline{91.00}   \\
%%%%%%%%%%%%%%%%%%%%%%%%%%%%%%%%%%%%%%%%%%%%%%%%%%%%%%%%%%%%%%%%%%%%%%%%%%%%%%%%%%%%%%%%%
\bottomrule

\end{tabular}
\end{table*}

\begin{table*}[t!]
\footnotesize
\caption{Table showing mean Intersection over Union (mIoU), Dice, Precision, Recall and accuracy scores for the SOTA and our proposed method on cataract surgery datasets. Best values are represented in \textbf{Bold} and second best \underline{underlined}. }\label{tab:cadisquant}
\begin{tabular}{lcccc|cccc}
% \textbf{EndoUDA (polyp)}&\multicolumn{2}{c}{\textbf{Source Modality}}& \multicolumn{1}{c}{\textbf{Target Modality}}\\
\toprule

% \textbf{CaDIS}   & &  & & &\multicolumn{4}{c}{\bf Cataract-1K (OOD)}\\
& \multicolumn{4}{c}{\bf CaDIS (IID)} & \multicolumn{4}{c}{\bf Cataract-1K (OOD)}\\

 \cline{2-5}  \cline{6-9} \\
\hline
Method & mIoU & mDSC &  Prec. & Rec.  &   mIoU &  mDSC & Prec. & Rec.  \\
\hline
% \hline
% \textbf{CholecSeg8K} &&& \\

Baseline \cite{chen2018encoder} & $73.64 \pm 0.76$  & $82.46 \pm 0.72$& 83.13 & \textbf{85.48} &   $29.92 \pm 0.75$ & $43.09 \pm 0.65$ & 31.54 & 65.23   \\

IBN-Net  \cite{pan2018two}& $72.90 \pm 0.82$ & $81.52 \pm 0.63$& 82.98 & 84.69   & $33.20 \pm 0.84$ & $48.56 \pm 0.73$ & 35.60 & 69.89  \\

RobustNet  \cite{choi2021robustnet}   & $73.06 \pm 0.84$ & $83.33 \pm 0.63$ & 83.26  & 84.40  & $41.58 \pm 0.72$ &  $52.63 \pm 0.70$ & 53.46 & \underline{77.51}    \\

SAMed \cite{SAMed} & $72.37 \pm 0.57$ & $82.85 \pm 0.42$ & 82.64 & 84.11   &  $41.84 \pm 0.60$   &   $47.38 \pm 0.52$ & 43.68 & 56.60     \\

SNR \cite{jin2021style} & $\underline{73.92} \pm \underline{0.78}$  & $\underline{83.44} \pm \underline{0.64}$ & \underline{83.53} & 84.63   & $40.37 \pm 0.51$ & $51.06 \pm 0.52$ & 49.37 & 73.53   \\

SAN-SAW \cite{peng2022semantic} & $68.67 \pm 0.60$  & $79.83 \pm 0.54$  & 81.30  & 79.84   & $39.97 \pm 0.67$  & $56.24 \pm 0.51$  & \textbf{62.34} & 54.30   \\

BlindNet \cite{ahn2024style} & $71.56 \pm 0.95$ & $81.92 \pm 0.76$ & 83.43 & 82.49  & $\underline{43.54} \pm \underline{0.63}$ & $\underline{53.87} \pm \underline{0.54}$ & 53.42 & 82.48 \\ 

RobustSurg (Ours)  &$\textbf{74.83} \pm \textbf{0.71}$&   $\textbf{84.78} \pm \textbf{0.50}$& \textbf{84.81} & \underline{85.39}& $\textbf{45.26} \pm \textbf{0.52}$ & $\textbf{58.34} \pm \textbf{0.41}$ & \underline{55.16} & \textbf{77.68} \\

\bottomrule
\end{tabular}
\end{table*}

\subsection{Discussion}

Domain generalisation is being extensively investigated in the natural image domain through various methods including style-content disentanglement frameworks~\citep{choi2021robustnet,zhao2022style,ahn2024style}, however, it is severely underexplored topic in the surgical domain. One important reason for this is the lack of availability of multicentre labeled datasets for the generalisability assessment of developed methods. To this end, we approach the domain generalisation problem from two perspectives, firstly, to address the important research gaps in the surgical domain having lack of diverse labeled datasets,  we introduce a multicentre labeled dataset with pixel-wise annotations for the generalisability evaluation of methods, and secondly we introduce style-content disentanglement based on instance whitening transformation~\citep{choi2021robustnet} from the natural domain to the surgical image domain by proposing RobustSurg method to effectively learn discriminative information to improve model generalisability. 

As reported by prior works~\citep{ahn2024style,jin2021style} and pictorially depicted in Fig.~\ref{fig:conceptual_flow}, previous DG methods predominantly using instance normalisation to suppress the image styles also remove useful features from the training data. We aim to alleviate this problem by adding style normalisation and restitution module~\citep{jin2021style} before the feature whitening unlike performing the feature whitening directly as reported in~\citep{choi2021robustnet}. Results on both in-domain and out-of-domain datasets as shown in Tables~\ref{tab1:results},~\ref{tab:endoudaquant},~\ref{tab:cadisquant} validate the effectiveness of our approach. Results on the Cholecystectomy surgery dataset in Table~\ref{tab1:results} shows that our method substantially improved upon both RobustNet and SNR in both in-domain and out-domain cases showing the effectiveness of using feature restitution before performing feature whitening. For instance, on the in-domain dataset RobustSurg outperformed RobustNet by 1.8\% and SNR by 1.2\% in terms of mean IoU score. On the generalisability tests on HeiCholeSeg, RobustSurg demonstrates a mIoU score improvement of 6.44\% and 5.54\% on center 1, and  10.45\%, 14.59\% on center 2 compared to RobustNet and SNR methods respectively. This performance gain implies that combining feature restitution with instance whitening actually encourages the model in learning discriminant features for better domain generalisation and the restitution stage serves to effectively remove style information as depicted in Fig.~\ref{fig:covariance}. Methods such as SHADE, SNR and BlindNet do well on HeiCholeSeg centre 1, but their performance degrades on centre 2. This implies that they struggle to deal with a higher image resolution. However, our method shows consistent performance in both centres demonstrating its robustness to the image resolutions. As RobustSurg essentially exploits the feature space in the feature extraction backbone, we also show in Table~\ref{tab-backbones} that our method consistently outperforms other SOTA on other important backbone networks such as MobileNetv2 and ShuffleNetv2 making RobustSurg a backbone-agnostic method. In Fig. ~\ref{fig:stat_test}, paired $t$-test shows RobustSurg achieved highly significant difference against other methods with narrower deviation on both in-domain and out-of-domain datasets demonstrating robust performance of our method. 

We observe the similar performance trends on other dataset setting, such as multi-modality EndoUDA dataset in Table~\ref{tab:endoudaquant}
 where training and test data come from a different modality. For example, RobustSurg surpassed RobustNet and SNR by 11.47\% and 9.7\% in terms of IoU score on EndoUDA polyp target set, and 21.79\%, and 19.26\% on EndoUDA Barrett's oesophagus target set in terms of IoU score. RobustSurg also demonstrates a consistent improvement in all evaluation metrics on cataract surgery datasets as shown in Table~\ref{tab:cadisquant}. For instance on CaDIS in-domain test set, RobustSurg obtained 2.4\% and 1.2\% higher IoU score compared to the RobustNet and SNR, and similarly on OOD Cataract-1K dataset, RobustSUrg outperformed RobustNet and SNR by 8.9\% and 12\% in terms of IoU score.

\section{Conclusion and future directions} \label{sec:conclusion}
In this work, we addressed the DG problem for surgical scene segmentation by addressing the weaknesses in the IN and WT. We propose a domain-invariant feature encoder where we integrate a restitution network to restore the discriminant information lost due to the instance normalisation, and ISW block to selectively suppress the style information and retain content useful for the model generalisability. We used DeepLabv3+ as the base model for semantic segmentation with ResNet50 as the backbone. We comprehensively evaluated our method on various datasets including laparoscopic and cataract surgery and an endoscopic data. We have also addressed the lack of multicentre data to evaluate model generalization by introducing a pixel-level annotated dataset named HeiCholeSeg, comprising of data from two different centres. Overall results on individual datasets shows the strong generalization performance of our method. To improve model performance on individual classes and address the performance degradadation on instrument classes due to vendor differences, we will work on individual class specific learning paradigms inside the segmentation decoder in future.

\section*{Acknowledgments}
The authors wish to acknowledge the Mexican Secretaría de Ciencia, Humanidades, Tecnología e Innovación (Secihti) for their support in terms of postgraduate scholarships in this project, and the Data Science Hub at Tecnologico de Monterrey for their support on this project.

This work has been supported by Azure Sponsorship credits granted by Microsoft's AI for Good Research Lab through the AI for Health program. The project was also supported by the French-Mexican ANUIES CONAHCYT Ecos Nord grant 322537, the WUN Research Development Fund 2024, the Crohn’s \& Colitis UK (M2023-5) and the Academy of Medical Sciences (SBF0010$\backslash$1191). {This work was also partially supported by the Worldwide Universities Network (WUN) Research Development Fund 2022 on ``Novel robust computer vision methods and synthetic datasets for minimally invasive surgery'' and Research Development Fund 2024 on ``Artificial Intelligence for Surgical Training Towards Safer and Effective Sub-mucosal Dissection''.}

{License agreement for HeiChole benchmark was published under}
{Creative Commons Attribution-NonCommercial-ShareAlike (CC BY-NC-SA)}

\bibliographystyle{cas-model2-names}

\bibliography{cas-refs}

%%%%%%%%%%%%%%%%%%%%%%%%%%%%%%%%%%%%%%%%%%%%%%%%%%%%%%%%%
%%%%%%%%%%%%%%%%%%%%%%%%%%%%%%%%%%%%%%%%%%%%%%%%%%%%%%%%%

\end{document}